\documentclass[10pt,twocolumn,letterpaper]{article}

\usepackage[pagenumbers]{cvpr} % To force page numbers, e.g. for an arXiv version    
\usepackage{circledsteps}

\usepackage{graphicx}
\usepackage{multirow}
\usepackage{bbding}
\usepackage{balance}
\usepackage{diagbox}
\usepackage{soul}
\usepackage{colortbl}

\definecolor{cvprblue}{rgb}{0.21,0.49,0.74}
\usepackage[pagebackref,breaklinks,colorlinks,citecolor=cvprblue]{hyperref}

\def\paperID{12328} % *** Enter the Paper ID here
\def\confName{CVPR}
\def\confYear{2024}

\title{Beyond Hallucinations: Enhancing LVLMs \\ through Hallucination-Aware Direct Preference Optimization}

\author{Zhiyuan Zhao\thanks{Equal contribution.}, Bin Wang\footnotemark[1], Linke Ouyang\footnotemark[1], Xiaoyi Dong, Jiaqi Wang, Conghui He\thanks{Corresponding author.}\\
Shanghai AI Laboratory\\
{\tt\small \{zhaozhiyuan, wangbin, ouyanglinke, dongxiaoyi, wangjiaqi, heconghui\}@pjlab.org.cn}
}

\begin{document}
\maketitle
\begin{abstract}

Multimodal large language models have made significant advancements in recent years, yet they still suffer from a common issue known as the ``hallucination problem", in which the models generate textual descriptions that inaccurately depict or entirely fabricate content from associated images.  This paper introduces a novel solution, Hallucination-Aware Direct Preference Optimization (HA-DPO), which reframes the hallucination problem as a preference selection task. The model is trained to favor the non-hallucinating response when presented with two responses of the same image (one accurate and one hallucinatory). Furthermore, this paper proposes an efficient pipeline for constructing positive~(non-hallucinatory) and negative~(hallucinatory) sample pairs, ensuring a high-quality, style-consistent dataset for robust preference learning. When applied to three mainstream multimodal models, HA-DPO significantly reduced hallucination issues and amplified the models' generalization capabilities. Notably, the MiniGPT-4 model, when enhanced with HA-DPO, demonstrated a substantial improvement: POPE accuracy rose from 51.13\% to 86.13\% (an absolute improvement of 35\%), and the MME score surged from 932.00 to 1326.46 (a relative improvement of 42.32\%). The codes, models, and datasets are made accessible at \url{https://opendatalab.github.io/HA-DPO.}

\end{abstract}    
\section{Introduction}
\label{sec:intro}

\begin{figure}
    \centering  
    \begin{subfigure}[b]{0.49\textwidth}
         \includegraphics[width=\textwidth]{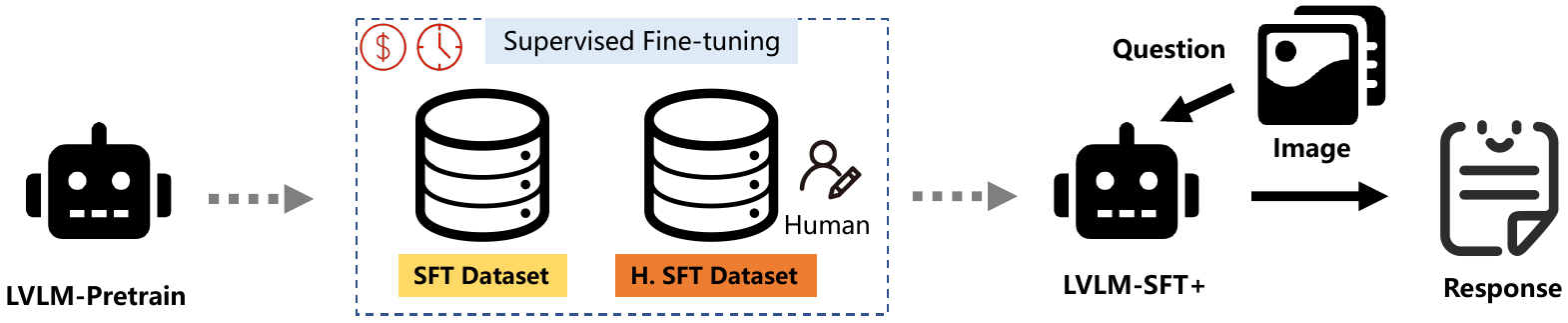}
            \caption{Hallucination Elimination with SFT}
            \label{fig:fig1_a}
    \end{subfigure}
    \begin{subfigure}[b]{0.49\textwidth}
         \includegraphics[width=\textwidth]{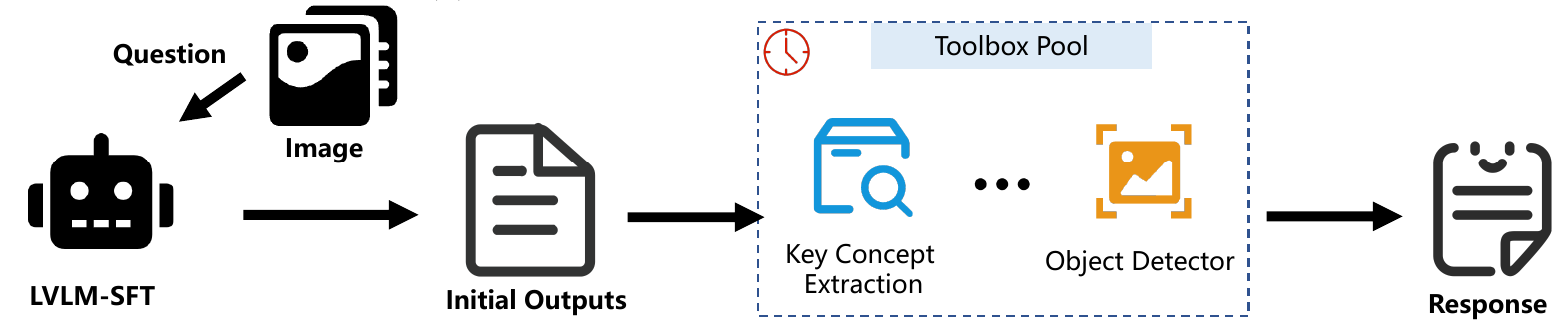}
            \caption{Hallucination Elimination with Post-hoc}
            \label{fig:fig1_b}
    \end{subfigure}
    \begin{subfigure}[b]{0.49\textwidth}
         \includegraphics[width=\textwidth]{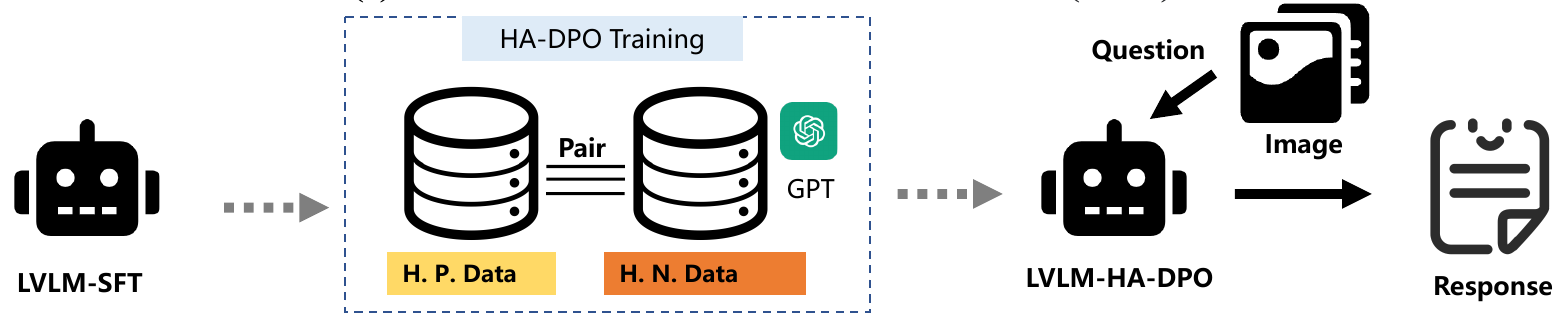}
            \caption{Hallucination Elimination with HA-DPO~(Ours)}
            \label{fig:fig1_c}
    \end{subfigure}
\vspace{-15pt}
    \caption{ Comparative diagram of strategies for hallucination elimination in LVLMs. `H.' represents hallucination, `H.P.' stands for positive samples without hallucination, and `H.N.' denotes negative samples with hallucination.}
\label{fig:fig1}
\vspace{-15pt}
\end{figure}

Large Vision-Language Models (LVLMs) have achieved substantial advancements in synchronizing visual and textual features, thanks to extensive unsupervised pre-training on image-text pairs and meticulous fine-tuning on high-quality data. This progress has endowed these models with the ability to follow multimodal instructions~\cite{DBLP:journals/corr/abs-2304-10592,DBLP:journals/corr/abs-2304-08485,DBLP:journals/corr/abs-2305-06500,DBLP:journals/corr/abs-2309-15112}, resulting in significant advancements in a variety of multimodal tasks. However, even the most advanced LVLMs are not immune to the pervasive issue of ``hallucination". This phenomenon, where the model's output includes fictitious information, such as non-existent objects or incorrect categories, attributes, or relationships, is a critical challenge~\cite{DBLP:journals/corr/abs-2305-10355,DBLP:journals/corr/abs-2306-14565,DBLP:journals/corr/abs-2308-06394,DBLP:journals/corr/abs-2309-14525,DBLP:journals/corr/abs-2310-00754,DBLP:journals/corr/abs-2310-01779}. These hallucinated details not only tarnish the user experience but can also mislead users, potentially triggering severe consequences. For instance, an inaccurate description in medical diagnostics could lead to a misdiagnosis, posing unforeseeable harm to patients.

Researchers have embarked on a series of studies to mitigate the hallucination issue in multimodal models. Current mainstream solutions can be broadly divided into two categories: the first, as shown in Figure~\ref{fig:fig1_a}, involves constructing abundant hallucination-free data and employing exhaustive Supervised Fine-Tuning (SFT) process to mitigate hallucination phenomenon~\cite{DBLP:journals/corr/abs-2306-14565,DBLP:journals/corr/abs-2305-06500,DBLP:journals/corr/abs-2309-15112}. Although SFT-based method reduces hallucination to a certain extent, this method demands high-quality data and comes with high annotation and training overheads. The second, as shown in Figure~\ref{fig:fig1_b}, treats the elimination of hallucination as a post-processing operation on model output and uses existing tools or expert models (such as tokenizers and detection models) to rectify hallucinated content~\cite{DBLP:journals/corr/abs-2310-00754}. This approach does not necessitate additional data and training, unless a certain customized model is required~\cite{DBLP:journals/corr/abs-2310-16045} for post-hoc hallucination processing. However, its effectiveness is constrained by existing resources, and the time cost increases with additional post-processing tools.

While the SFT method is beneficial, it lacks flexibility for algorithm optimization and custom hallucination handling, and it demands substantial resources for data construction and training. Conversely, the Post-Hoc method's performance wholly depends on the existing toolset. In contrast, this paper perceives hallucination elimination as a model preference, biasing the model towards hallucination-free output. Following this approach, we decouple the task of multimodal hallucination elimination into a preference optimization problem. Through this preference learning strategy, we anticipate that multimodal models will exhibit a preference constraint during training, leading to a bias toward hallucination-free output. RLHF~\cite{DBLP:journals/corr/abs-2204-05862} and DPO~\cite{DBLP:journals/corr/abs-2305-18290} are two effective strategies proposed to address the preference issue in large language models. Given the lightweight nature and efficient training of the DPO, this paper extends DPO to propose a multimodal DPO strategy sensitive to hallucinations~(\textbf{H}allucination-\textbf{A}ware \textbf{D}irect \textbf{P}reference \textbf{O}ptimization, termed \textbf{HA-DPO}), thereby customizing and enhancing the hallucination elimination capability of multimodal models.

In scrutinizing current methodologies for multimodal hallucination evaluation, we've identified significant shortcomings in existing assessment systems~\cite{DBLP:conf/emnlp/RohrbachHBDS18,DBLP:journals/corr/abs-2305-10355,DBLP:journals/corr/abs-2308-15126}. Firstly, these systems limit categories to predefined ones, leading to the erroneous marking of unrecognized content as hallucinations. Additionally, they predominantly focus on specific areas, such as the existence and attributes of objects, neglecting a wider range of hallucinations. To address these issues, we introduce the Sentence-level Hallucination Ratio (\textbf{SHR}), a comprehensive and intuitive benchmark. Unconfined by fixed categories and scopes, the \textbf{SHR} offers a broad, fine-grained, and quantitative measurement for multimodal hallucinations.

In summary, our contributions are as follows:
\begin{enumerate}
    \item We propose the HA-DPO strategy, a novel paradigm specifically designed to overcome hallucinations in large multimodal models, which is demonstrated in Figure~\ref{fig:fig1_c}. Additionally, we develop a method for constructing style-consistent hallucination sample pairs, ensuring the stability of HA-DPO training.

    \item We introduce the Sentence-level Hallucination Ratio (SHR), an intuitive and comprehensive metric for assessing hallucinations in LVLMs.

    \item Through extensive experiments on prevalent models, we demonstrate the effectiveness of our approach, showing a marked reduction in hallucinations and a notable enhancement in the general performance of the models.
\end{enumerate}

\section{Related Work}
\label{sec:related_work}

\subsection{Hallucination in LLMs}

Hallucinations in Large Language Models (LLMs) refer to a phenomenon where the model generates responses that conflict with known facts \cite{DBLP:journals/corr/abs-2307-11019, DBLP:conf/acl/LinHE22, DBLP:journals/corr/abs-2309-01219}. Early research has indicated that hallucinations are often attributed to noise in the pre-training data~\cite{DBLP:journals/corr/abs-2305-14552, DBLP:conf/nips/Ouyang0JAWMZASR22} and a lack of prior knowledge during supervised fine-tuning (SFT)~\cite{DBLP:conf/nips/Ouyang0JAWMZASR22}. To address these issues, Falcon~\cite{DBLP:journals/corr/abs-2306-01116} has developed a powerful strategy for cleaning internet data. LLaMA2~\cite{DBLP:journals/corr/abs-2307-09288} reduces data noise by increasing the sampling ratio of high-quality data sources. As for hallucination mitigation during SFT, LIMA~\cite{DBLP:journals/corr/abs-2305-11206} suggests reducing SFT data volume for enhanced performance. Reinforcement Learning from Human Feedback~(RLHF) is another strong method for mitigating hallucinations. For instance, by employing RLHF, GPT4 enhances the accuracy of TruthfulQA from 30\% to 60\%~\cite{DBLP:journals/corr/abs-2303-08774}. 

 \begin{figure*}[t]
    \centering
    \includegraphics[width=\textwidth]{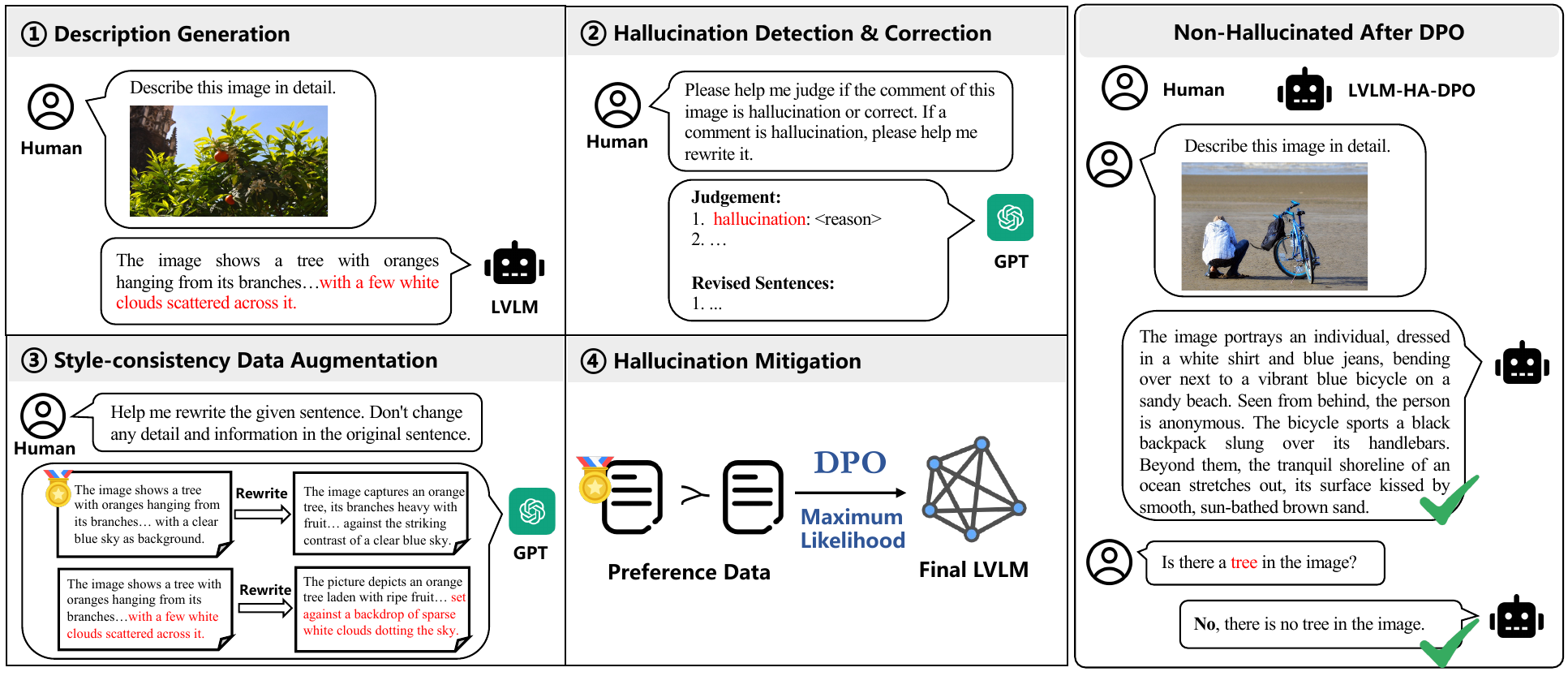}
    \vspace{-18pt}
    \caption{Our proposed hallucination mitigation process involves four steps: \textbf{(1) Description Generation}, where the LVLM is tasked with a detailed image description; \textbf{(2) Hallucination Detection and Correction}, GPT-4 identifies and corrects hallucinations in model responses using rich annotations; \textbf{(3) Style-consistency Data Augmentation}, GPT-4 rewrites samples to maintain style consistency; and \textbf{(4) Hallucination Mitigation}, style-consistent data is gathered for DPO training.}
    \label{fig:fig2}
\end{figure*}

\subsection{Hallucination in LVLMs}
The hallucination issue in LVLMs, exacerbated by language and image modal space misalignment, has recently become the focus of numerous studies. POPE~\cite{DBLP:journals/corr/abs-2305-10355} identifies the object existence hallucination issue in LVLM, illustrating its severity. LRV~\cite{DBLP:journals/corr/abs-2306-14565} creates abundant questions with rich annotations and GPT4-assisted evaluation. However, existing benchmarks fail to cover a sufficient range of object categories and types of hallucinations.

%NOPE~\cite{DBLP:journals/corr/abs-2310-05338} examined 10 advanced LVLMs for their capacity to detect non-existence of objects in visual questions, revealing that all are susceptible to object hallucination. 

In terms of mitigating hallucination, most methods have focused on improving the quality of the SFT data. VIGC~\cite{DBLP:journals/corr/abs-2308-12714} introduces a module for visual instruction correction to minimize long sequence generation hallucinations. LRV~\cite{DBLP:journals/corr/abs-2306-14565} and InstructBLIP~\cite{DBLP:journals/corr/abs-2305-06500} mitigate hallucinations by constructing substantial and diverse SFT data. LLaVA-RLHF~\cite{DBLP:journals/corr/abs-2309-14525} is the first to apply RLHF in hallucination mitigation of LVLM. Gunjal~\textit{et al.} utilize DPO~(Direct Preference Optimization)~\cite{DBLP:journals/corr/abs-2305-18290} for hallucination reduction, but use only dis-preferred data~\cite{DBLP:journals/corr/abs-2305-18290}. However, their approach risked over-exploitation due to lack of preferred data, complicating effective strategy learning~\cite{DBLP:conf/nips/ChristianoLBMLA17,DBLP:conf/intellisys/MousaviSH16,berger2014exploration,DBLP:journals/jmlr/OsbandRRW19}.  In contrast, our framework integrates high-quality, human-free, style-consistent data into DPO, enhancing data quality and RL strategy beyond previous methods.

\subsection{Human-preference Learning}
Human-preference learning is a crucial approach in constructing effective, safe, and trustworthy models.~\cite{DBLP:journals/corr/abs-2303-08774,DBLP:journals/corr/abs-2307-09288,DBLP:conf/nips/Ouyang0JAWMZASR22}. RLHF~\cite{DBLP:conf/nips/Ouyang0JAWMZASR22,DBLP:journals/corr/abs-1909-08593,DBLP:conf/icml/MacGlashanHLPWR17} has been the most successful implementation in learning human preference. It constructs a reward model based on human preferences and optimizes the policy model guided by feedback from the reward model.~\cite{DBLP:journals/corr/SchulmanWDRK17}. Similarly, InstructGPT exploits human preference data for RL optimization, building upon GPT3~\cite{DBLP:conf/nips/BrownMRSKDNSSAA20}. Bai~\textit{et al.}~\cite{DBLP:journals/corr/abs-2204-05862} aim to enhance human-preference learning with updated RL policies.  Recently, DPO proposes a method~\cite{DBLP:journals/corr/abs-2305-18290} that bypasses learning the reward model and directly learns policies, resulting in a much simpler approach with superior performance compared to PPO. The current state-of-the-art AI agent, GPT4, also utilizes human-preference alignment via extensive data processing during its construction~\cite{DBLP:journals/corr/abs-2303-08774}.
\section{Our Method}
\label{sec:our_method}

% In this work, our proposed HA-DPO extends DPO to multimodal domains to improve the authenticity of the multimodal model. As illustrated in Figure~\ref{fig:fig2}, Our proposed HA-DPO consists of four steps: (1)~Description generation, (2)~Hallucination detection and correction, (3)~Style-consistent data augmentation, and (4) Hallucination Mitigation. The goal of steps 1 to 3 is to construct a style-consistent hallucination-aware dataset. This dataset is subsequently utilized for the hallucination mitigation in step 4. The models trained using HA-DPO demonstrate a significant reduction in hallucination phenomena in tasks involving detailed descriptions and dialogues.

In this work, we introduce a novel method, Hallucination-Aware Direct Preference Optimization (HA-DPO), designed to constrain the model's preference toward outputs that are devoid of hallucinations. The primary objective is to promote non-hallucinatory outputs. Considering the inherent complexity in data construction and model training processes when utilizing classical RLHF methods to constrain model preferences, we opt for Direct Preference Optimization (DPO). DPO, being a simpler, more efficient, and RL-free method, serves as our foundational strategy. We extend this strategy to the multimodal domain with the intention of eliminating hallucinations, thereby enhancing the authenticity and precision of multimodal model outputs.

As illustrated in Figure~\ref{fig:fig2}, we construct a style-consistent hallucination dataset through three primary steps: {\footnotesize{\Circled{1}}}~description generation, {\footnotesize{\Circled{2}}}~hallucination detection and correction, and {\footnotesize{\Circled{3}}}~style-consistent data augmentation. This dataset is subsequently utilized for the training of the HA-DPO model. The models trained using this method demonstrate a significant reduction in hallucination phenomena in tasks involving detailed descriptions and dialogues.

\subsection{Multimodal Hallucination-Aware DPO}

We formulate the elimination of hallucinations as a preference selection problem, where, in conjunction with a style-consistent dataset, preference learning is conducted. This encourages the model to favor non-hallucinatory positive response $y_{pos}$ and reject hallucinatory negative response $y_{neg}$. During the human-preference learning stage, a reward model, denoted as $\hat{r}$, is trained to ensure the model's propensity for outputting non-hallucinated responses $y_{pos}$ and rejecting hallucinated responses $y_{neg}$. This reward model is capable of assigning scores to different responses $y$, thereby accurately reflecting human preferences. Once the reward model $r$ is obtained, it is used to provide feedback for guiding an additional fine-tuning phase, which learns a policy model $\pi_{\theta}$ guided by human preference. 

Building on the DPO framework, HA-DPO avoids the need for implicit reward model learning, instead directly optimizing the policy model $\pi_{ref}$ as follows:

% During the human-preference learning phase, a reward model $\hat{r}$ is trained to favor non-hallucinated response $y_{pos}$ and reject hallucinated responses $y_{neg}$. effectively scoring responses $y$ according to human preferences on hallucination. Upon obtaining the reward model $\hat{r}$, it guides an additional fine-tuning phase, teaching a policy model $\pi_{\theta}$ based on human preference. Building upon the DPO framework, HA-DPO circumvents the need for implicit reward model learning and directly optimizes the policy model $\pi_{ref}$ as follows:
\vspace{-5pt}
\begin{equation}
\begin{aligned}
    L_{dpo}(\pi_{\theta};\pi_{ref}) = &-E_{(x_T,x_I,y_{pos},y_{neg}) \sim D} \\
    &\biggl[ \log \sigma (\beta log \frac{\pi_{\theta}(y_{pos}|[x_T,x_I])}{\pi_{ref}(y_{pos}|[x_T,x_I])} \\
    & - \beta \log \frac{\pi_{\theta}(y_{neg}|[x_T,x_I])}{\pi_{ref}(y_{neg}|[x_T,x_I])}) \biggr]
    \label{eq1}
\end{aligned}
\end{equation}
%\noindent where $x_T$ and $x_I$ refer to text and image prompts, $[~]$ represents feature concatenation, $\pi_{ref}$ and $\pi_{\theta}$ are reference and policy models, respectively. $D$ denotes the style-consistent dataset, and $\log \sigma$ denotes the log-sigmoid function. This objective concurrently trains reward and policy models, biasing the reward model towards favoring positive response $y_{pos}$ selection and the negative response $y_{neg}$ rejection. Instead of explicit reward model training, Eq.~\ref{eq1} directly optimizes the policy model $\pi_{ref}$, with the reward model $r$ implicitly represented by:

\noindent where $x_T$ and $x_I$ denote the text and image prompts, respectively, while $[~]$ signifies feature concatenation. We refer to $\pi_{ref}$ and $\pi_{theta}$ as the reference and policy models, respectively. $D$ represents the style-consistent dataset, and $\log \sigma$ stands for the log-sigmoid function. This objective function is designed to train the reward and policy models concurrently, skewing the reward model to favor positive responses $y_{pos}$ and reject negative responses $y_{neg}$. Instead of training the reward model explicitly, Eq.~\ref{eq1} optimizes the policy model $\pi_{ref}$ directly, with the reward model $r$ represented implicitly as follows:
\vspace{-5pt}
\begin{equation}
    \hat{r}(x_T,x_I,y) =  \beta log \frac{\pi_{\theta}(y_{pos}|[x_T,x_I])}{\pi_{ref}(y_{pos}|[x_T,x_I])}
    \label{eq2}
\end{equation}
Building on Eq.~\ref{eq2}, we formulate Eq.~\ref{eq1} to maximize the reward margin $\hat{r} (x_T,x_I,y_{pos}) - \hat{r} (x_T,x_I,y_{neg})$. This approach effectively amplifies the log-likelihood of the positive, non-hallucinated sample $y_{pos}$, while diminishing that of the negative, hallucinated sample $y_{neg}$. Consequently, the model is steered to favor non-hallucinated over hallucinated samples. To ensure training stability, we incorporate an auxiliary causal language modeling task into the preference learning process, drawing inspiration from InstructGPT~\cite{DBLP:conf/nips/Ouyang0JAWMZASR22}. The auxiliary task is defined as follows:

%Eq.~\ref{eq2} is the representation of implicit reward model representation. Objective~\ref{eq1} maximizes the reward margin $\hat{r} (x_T,x_I,y_{pos}) - \hat{r} (x_T,x_I,y_{neg})$ and increases the log-likelihood of the positive non-hallucinated sample $y_{pos}$ and decreases that of the negative hallucinated sample $y_{neg}$, thereby guiding the model to favor non-hallucinated samples over hallucinated ones. {Besides, inspired by InstructGPT~\cite{DBLP:conf/nips/Ouyang0JAWMZASR22}, to preserve training stability, we incorporate an auxiliary causal language modeling task into preference learning process:

\begin{equation}
    L_{aux} = -\sum \log P(y | x_{P}; \pi_{\theta}), \{x_P, y\} \sim D_{sft}
    \label{eq_aux}
\end{equation}

\begin{figure}[t]
    \centering
    \includegraphics[width=0.49\textwidth]{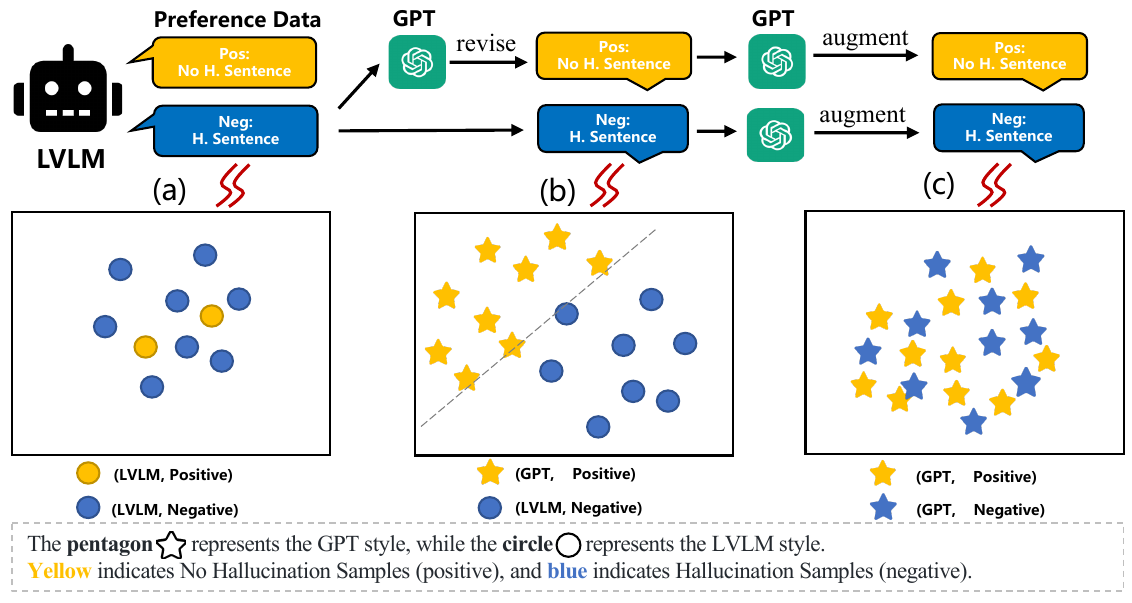}
    \vspace{-18pt}
    \caption{Style Consistency Analysis for Hallucination Dataset. }
    %(a) The distribution of positiva-negative data generated by LVLMs. The positive samples are rare, making it difficult to collect. (b) The distribution of negative samples generated from LVLMs and its corresponding positive samples corrected by GPT. There is an significant differences in distribution between the positive and negative samples, meaning they originate from different stylistic models. (c) The distribution of positive-negativa data rewrited by GPT. It illustrated that style-consistency positive and negative data shows little misalignment.
    \label{fig:fig3}
\vspace{-22pt}
\end{figure}
\noindent where $x_P$ represents the prompt and $y$ corresponds to the associated response. $D_{sft}$ denotes the data utilized during the SFT training phase, and $\pi_{\theta}$ refers to the policy model. Eq.~\ref{eq_aux} integrates the supervised fine-tuning gradient into the preference learning process, which serves to mitigate potential performance regression and ensure stable training. During HA-DPO training, we minimize an objective where $\lambda$ balances preference learning loss and auxiliary language modeling loss:

\vspace{-5pt}
\begin{equation}
    L = L_{dpo} + \lambda L_{aux}
    \label{eq_all}
\end{equation}

\subsection{Dataset Construction with Style Consistency}

%The HA-DPO method proposed in this paper requires pairs of positive and negative samples. Specifically, for a given image, the positive output is a detailed description $y_{pos}$ without hallucinations. In contrast, the negative output is a detailed description $y_{neg}$ that includes hallucinations.

\subsubsection{Data Source} This paper uses the Visual Genome (VG) dataset~\cite{DBLP:journals/ijcv/KrishnaZGJHKCKL17} to construct hallucinated samples $y_{neg}$ and non-hallucinated $y_{pos}$. The VG dataset contains abundant annotated information. Each image includes multiple region bounding boxes, each corresponding to a detailed description. These annotations can adequately cover various detailed information related to the image: diverse objectives, attributes, relationships, etc., unrestricted by specific categories and scopes.
\vspace{-5pt}
\subsubsection{Generation of Hallucination Sample Pairs.} As shown in Figure~\ref{fig:fig2}, based on Visual Genome images with detailed annotation information, we propose the following data construction process:

\noindent\textbf{Description Generation.} We randomly select images from the VG dataset and use the LVLM to generate corresponding detailed descriptions.

\noindent\textbf{GPT-4 Hallucination Detection and Correction.} Next, input the model-generated description and all the annotation information of the original image into GPT-4 and provide a detailed prompt template to enable GPT-4 to check whether there are hallucinations in the generated description. If hallucinations exist, a corrected description without hallucinations needs to be provided. This way, we can obtain the positive and negative responses corresponding to an image. In fact, hallucinations almost always occur when a multimodal model provides a detailed image description.

\begin{figure}[t]
    \centering
    \setcounter{subfigure}{0}
    \begin{subfigure}[b]{0.9\linewidth}
        \includegraphics[width=0.495\textwidth]{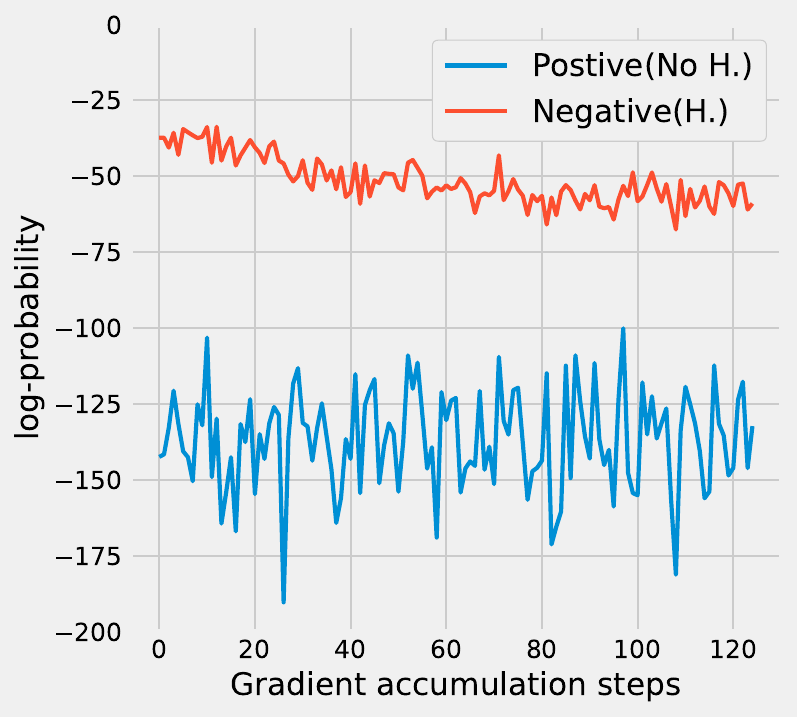}
        \includegraphics[width=0.478\textwidth]{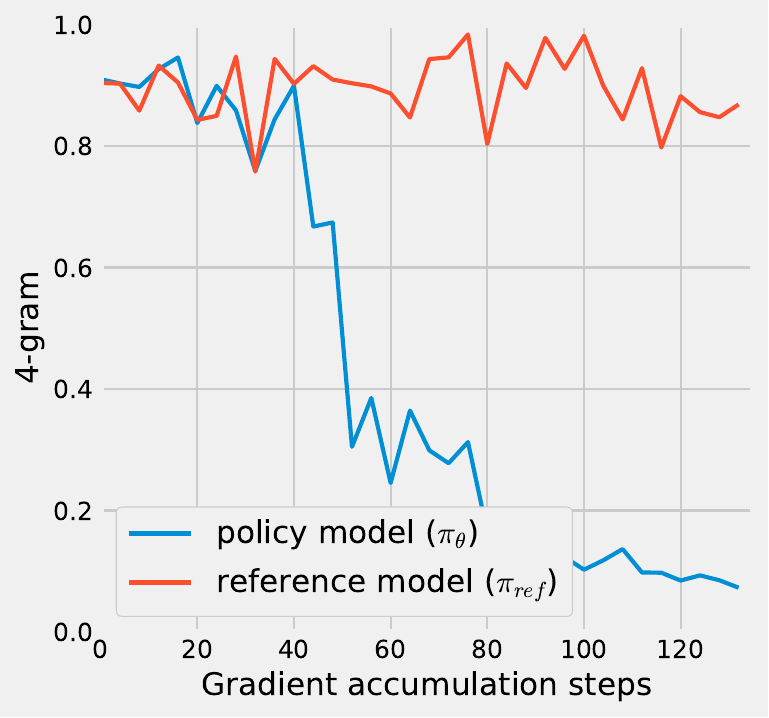}
        \caption{w/o style consistency control. Left: log-likelihood distribution; Right: N-grams comparison between reference and policy model. }
    \label{fig:fig4_a}
    \end{subfigure}

    \begin{subfigure}[b]{0.9\linewidth}
        \includegraphics[width=0.495\textwidth]{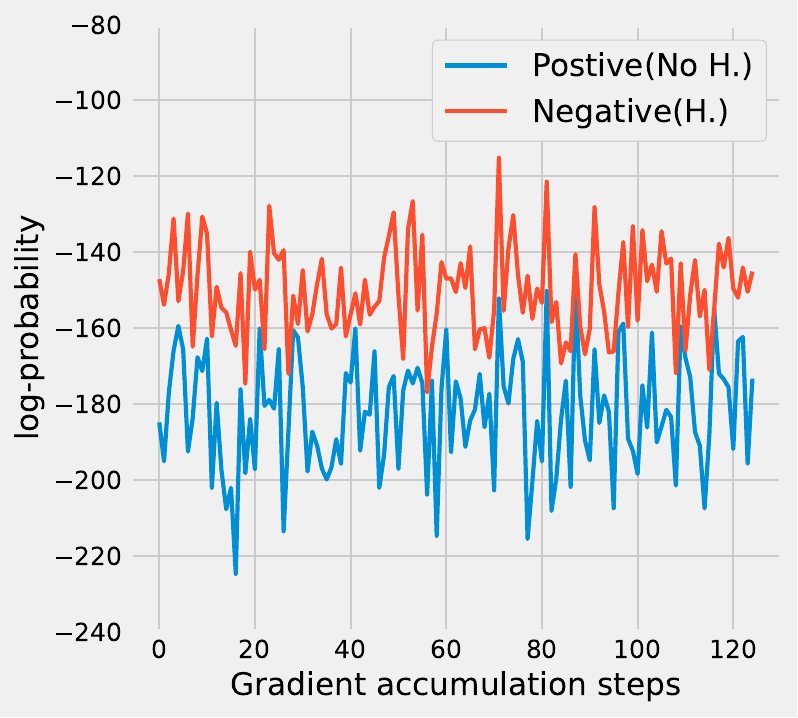}
        \includegraphics[width=0.478\textwidth]{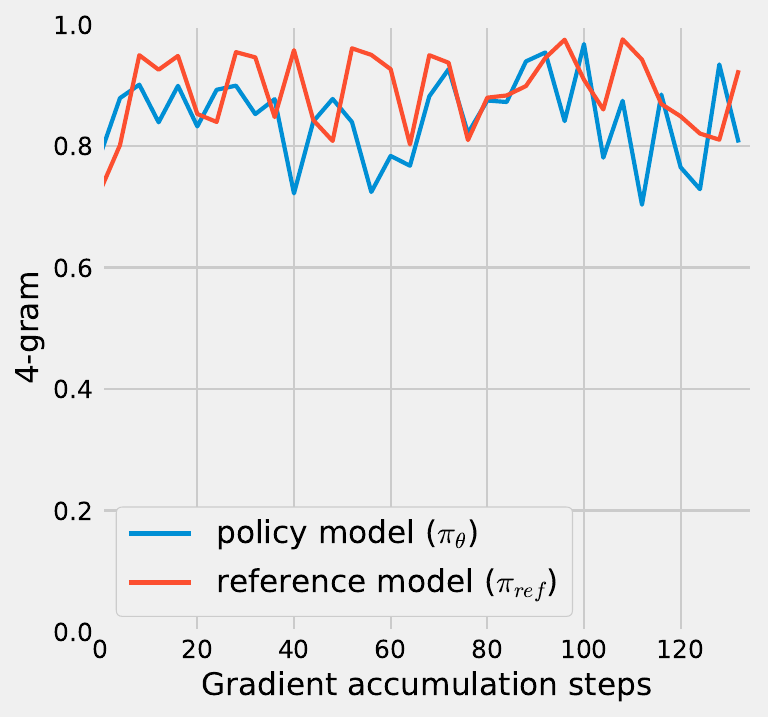}
        \caption{w/ style consistency control. Left: log-likelihood distribution; Right: N-grams comparison between reference and policy model. }
    \label{fig:fig4_b}
    \end{subfigure}
    
    \caption{Quantative analysis on style-consistent control. }
    \label{fig:fig4}
\vspace{-5pt}
\end{figure}

\begin{figure*}[t]
    \centering
    \includegraphics[width=1.0\textwidth]{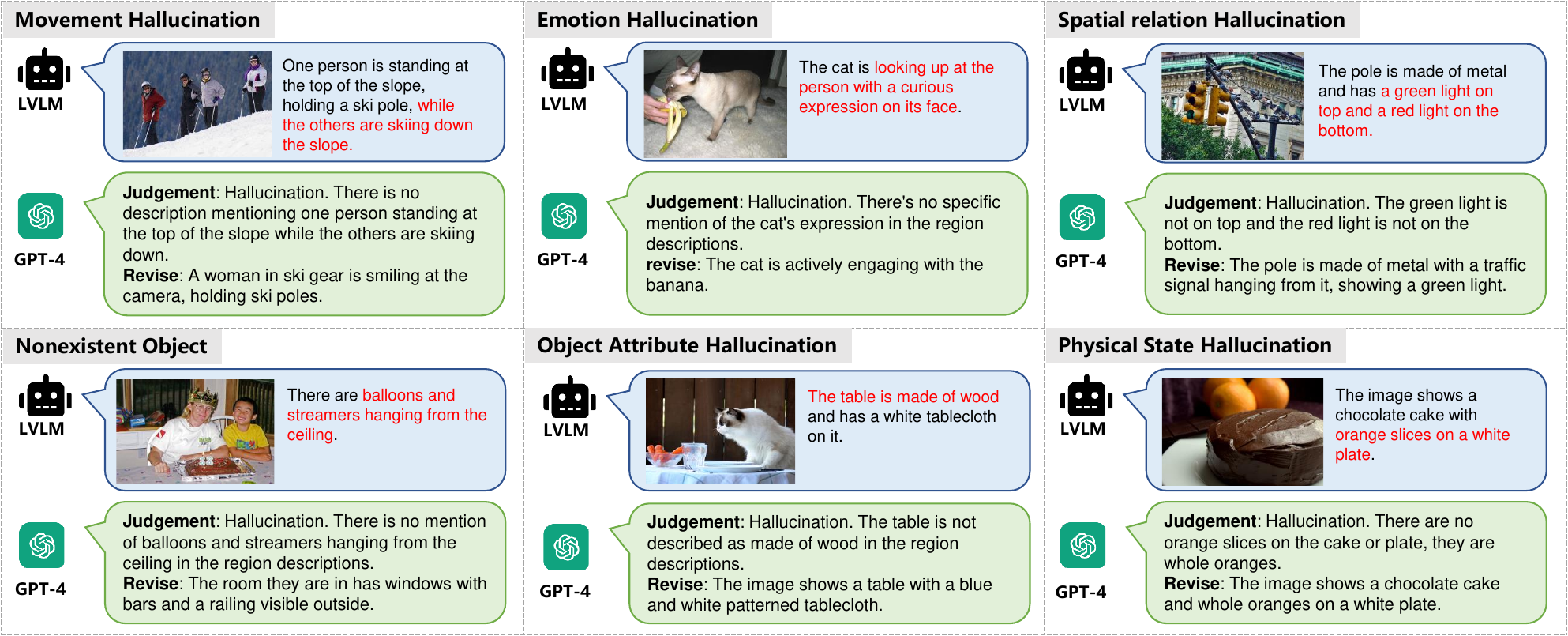}
    \caption{SHR evaluation covers diverse types of hallucinations.}
    \label{fig:shr}
\vspace{-15pt}
\end{figure*}

\noindent\textbf{Style-consistent Data Augmentation.} %To ensure style consistency between positive and negative sample sentences and obtain more samples, we use GPT-4 to rewrite the positive and negative samples obtained in the previous step, ensuring that the positivity and negativity remain unchanged. Besides, we further augment positive and negative data into question-answering format. Specifically, we let GPT-4 convert descriptive positive-negative data into one question with positive-negative answer pairs. Then the question is provided to LVLM and positive-negative responses are sampled according to previously given answers. In HA-DPO training, the descriptive positive-negative data and question-answering positive-negative data form all the preference learning data.
% To ensure style consistency between positive and negative sample sentences and obtain more samples, we use GPT-4 to rewrite the earlier positive and negative samples, preserving their positivity and negativity. Additionally, we augment these data into a question-answering format. Specifically, GPT-4 converts descriptive data into a question with positive-negative answer pairs, which LVLM is then provided with. The responses are sampled based on previous answers. In HA-DPO training, both descriptive and question-answering data constitute all preference learning data.
To ensure style consistency and amplify our sample pool, we use GPT-4 to rewrite earlier positive and negative samples, preserving their polarity. These data are further augmented into a question-answer format. Specifically, GPT-4 transforms descriptive data into questions coupled with positive-negative answer pairs, which are then fed into LVLM. Responses are sampled based on prior answers. In HA-DPO training, the preference learning data consists of both descriptive and question-answer data. Consequently, high-quality positive $(x, y_{pos})$ and negative $(x, y_{neg})$ sample pairs for the HA-DPO training process have been successfully constructed.

\subsection{Style Consistency Analysis}

High-quality and style-consistent data are crucial for the effective training of HA-DPO, as demonstrated in Figure~\ref{fig:fig3}-(a), which illustrates our initial generation of detailed image descriptions using a multimodal model that yields a small proportion of responses free from hallucinations. However, HA-DPO training necessitates a balance of positive and negative samples, $(x, y_{pos})$ and $(x, y_{neg})$, as indicated in Equation~\ref{eq1}. To achieve this, we utilize GPT-4 to refine hallucinated responses and produce positive responses without hallucinations, as shown in Figure~\ref{fig:fig3}-(b). Despite this, the observed distribution difference between the non-hallucinatory positive samples and hallucinatory negative samples does not result from the presence of hallucination. Instead, it arises from the stylistic variations between the direct outputs of the VLM and the refined outputs from GPT-4. This could potentially mislead the DPO into learning stylistic differences instead of distinguishing between hallucination-free and hallucination-prone responses. In order to address this, we utilize GPT-4 to augment the initial positive and negative samples in Figure~\ref{fig:fig3}-(b), as depicted in Figure~\ref{fig:fig3}-(c). At this point, both the hallucinatory negative samples and non-hallucinatory positive samples adopt the style of GPT-4, ensuring style consistency. Consequently, this enables the HA-DPO to learn a true preference for non-hallucination, thereby eliminating model hallucinations.

% Meanwhile, we examine the log-likelihood $\log \pi(y|x_T,x_I)$ distribution for positive and negative samples in Figure~\ref{fig:fig4}, where Figure~\ref{fig:fig4_a} and Figure~\ref{fig:fig4_b} shows data distribution without and with style-consistency. Directly revising negative responses using GPT-4 can lead to misalignment in data distributions, leading to training instability. However, this misalignment can be mitigated by style-consistency augmentation. Reward margin $r_{\theta}(x_T,x_I,y_{pos}) - r_{\theta}(x_T,x_I,y_{neg})$ is depicted in Figure~\ref{fig:fig4_c}. The absence of style consistency makes the reward model learn to distinguish samples based on style differences instead of hallucination, leading to training failure.

To intuitively understand the impact of style-consistency on HA-DPO training, we analyzed its effects on data distribution and sentence fluency during the training process. Prior to applying style-consistency alignment, as shown on the left side of Figure~\ref{fig:fig4_a}, there is a significant distribution difference between non-hallucination negative samples and hallucination positive samples. Concurrently, the right side of Figure~\ref{fig:fig4_a} illustrates a rapid decline in sentence fluency as training progresses, with the model losing its question-answering capability after a certain training period, manifested as repetitive output of the same word or sentence. However, after style-consistency alignment, as shown on the left side of Figure~\ref{fig:fig4_b}, the distributions of both types of samples are successfully pulled into the same feature space. Under this condition, sentence fluency is not affected as training progresses, and the model trained ultimately possesses the ability to eliminate hallucinations.%~(as shown in the right side of Figure~\ref{fig:fig4_b}).

Moreover, we corroborate, from the perspective of gradient optimization of objective~\ref{eq1}, that a misalignment in the distribution of positive and negative samples can induce instability during training. By deriving the optimization objective from objective~\ref{eq1}, we obtain:
\vspace{-5pt}
\begin{equation}
\begin{aligned}
& \nabla L  (\pi_{\theta} ;\pi_{ref}) = - \beta {E}_{x_T,x_I,y_{pos},y_{neg}} \\ 
&\sim \mathcal{D} \biggl[ \underbrace{\sigma(\hat{r}_{\theta}(x_T,x_I,y_{neg}) - \hat{r}_{\theta}(x_T,x_I,y_{pos}))}_{\text{weighting coefficient}} \\
&- \underbrace{[\nabla_{\theta} \log \pi (y_{pos}|[x_T,x_I]) - \nabla_{\theta} \log \pi (y_{neg}|[x_T,x_I)]]}_{\text{dominate gradient when $y_{pos}$ and $y_{neg}$ are misaligned}} \biggr]
\label{eq3}
\end{aligned}
\end{equation}
where the optimization gradient consists of two components. The first, represented by the reward model $\hat{r}$, serves to realign the model when the reward estimates deviate. The second component aims to enhance the log-likelihood of $y_{pos}$ while diminishing that of $y_{neg}$. As empirically demonstrated by \cite{DBLP:journals/corr/abs-2305-18290}, the weight factor of the reward model plays a pivotal role in maintaining the stability of training. As suggested by Eq.~\ref{eq3}, once the distribution of positive and negative samples ($y_{pos}$ and $y_{neg}$) becomes misaligned, this misalignment predominates the gradient, thereby destabilizing training despite the presence of the weighting factor. Further analysis can be found in Sec~\ref{supp:sec10}.

%The optimization gradient comprises two components. The first one, represented by the reward model $\hat{r}$, realigns the model when the reward estimates are wrong. The second part aims to increase the log-likelihood of $y_{pos}$ and decrease the log-likelihood of $y_{neg}$. \cite{DBLP:journals/corr/abs-2305-18290} empirically demonstrated that the weight factor of the reward model is very crucial for the stability of training. As seen from Eq.~\ref{eq3}, once the distribution of positive and negative samples~($y_{pos}$ and $y_{neg}$) are misaligned, this misalignment dominates the gradient, destabilizing training despite the weighting factor. More analysis can be found in Sec~\ref{supp:sec10}.

\section{{Sentencel-level Hallucination Ratio~(SHR)}}
\label{sec:shr}

\subsection{Benchmark Details}
% Currently, there are no accurate and wide-coverage hallucination evaluation benchmarks for LVLM. POPE~\cite{DBLP:journals/corr/abs-2305-10355}, which is the current most popular LVLM hallucination evaluation benchmark, lacks evaluation breadth as it only covers 80 COCO classes and excludes only object existence hallucination. In response to the limitations of the POPE, we propose a new evaluation metric, termed ``Sentence-level Hallucination Ratio" (SHR), aimed at quantifying hallucination extent at the sentence level in multimodal AI models, which can generate detailed, multiple sentences descriptions for an image. 

%At present, there are no benchmarks available that provide both accurate and comprehensive hallucination evaluation for LVLM. 

The currently prevalent LVLM hallucination evaluation benchmark, POPE~\cite{DBLP:journals/corr/abs-2305-10355}, restricts its evaluation scope to a limited number of target categories, neglecting a broader range of categories, attributes, emotions, and other elements. To overcome the limitations posed by POPE, we introduce a new evaluation metric, the SHR (Sentence-level Hallucination Ratio). 

The SHR aims to provide a comprehensive and broad measurement of whether the output of LVLMs contains hallucinations. The hallucination check is not confined to categories or attributes, but encompasses all textual descriptions that do not match the image content, thereby offering a quantifiable metric at the sentence level. The formal definition of the SHR is :
\begin{equation}
\begin{aligned}
SHR = \frac{\sum_{i=1}^{N} h_i}{\sum_{i=1}^{N} s_i} 
\end{aligned}
\end{equation}
where $N$ is the total number of images, $s_i$ and $h_i$ denote the number of hallucinated sentences and all sentences in the response, respectively. Specifically, we randomly selected 200 images from VG as the validation set. The determination of $h_i$ is made by GPT-4 based on the model's outputs and the corresponding annotations for the current image.

\subsection{Benchmark Advantages}
% Our proposed SHR outperforms existing benchmarks~(like POPE~\cite{DBLP:journals/corr/abs-2305-10355} and HaELM~\cite{DBLP:journals/corr/abs-2308-15126}) in the following ways: (1) Evaluation is accurate, benefited with manually annotated factual information, which helps improve accuracy of the GPT-4 judgment and leads to a judgment accuracy of about 95\%~(details shown in Sec~\ref{supp:sec9}). (2) It ensures openness of object categories, allowing evaluation of unlimited object types within Visual Genome (VG) images, not restricting to a few~(such as COCO 80 categories). (3) SHR accommodates different hallucination types. It considers any description conflicting with image content as a hallucination, therefore covering various hallucinations including nonexistent objects, emotional hallucinations, object attributes hallucinations, movement hallucinations, and more (as depicted in Figure~\ref{fig:shr}).
Our proposed SHR offers a more compelling evaluation than its counterparts, such as POPE and HaELM. It excels in three key areas: (1) \textit{Reliability}: SHR relies on manually annotated factual information to enhance GPT-4's judgment accuracy, achieving an impressive accuracy rate of approximately 95\% (see Section~\ref{supp:sec9} for more details). (2) \textit{Universality}: unlike benchmarks limited to a select few categories (such as COCO's 80), SHR accommodates an unlimited number of object types within VG images. 
(3) \textit{Comprehensiveness}: SHR can recognize a wide spectrum of hallucinations, tagging any description that contradicts image content as a hallucination. This includes hallucinations involving nonexistent objects, emotions, attributes, movements, and more (see Figure~\ref{fig:shr} for illustrations).

\section{Experiments}
\label{sec:experiment}

\subsection{Data and Evaluation}
\label{sec:data}

\noindent \textbf{{Training Data:}}
Based on the VG dataset~\cite{DBLP:journals/ijcv/KrishnaZGJHKCKL17}, we filtered out images that had either too many or too few objects, or those with insufficient annotation information, and randomly selected 2K images. The hallucination dataset was built using GPT-4 in three rewrites, yielding 2K images with 6K non-hallucinatory and 6K hallucinatory responses. With the conversion of descriptive data into a question-answer format, we added about 10K data pairs for training, summing up to ~16K total data used.

\begin{table}[t]
    \centering
    \resizebox{0.45\textwidth}{!}{%
    \begin{tabular}{c|c|c|c|c|c}
        \toprule
        \textbf{$\beta$} & \textbf{SHR~$\downarrow$} & \textbf{1-gram} & \textbf{2-gram} & \textbf{3-gram} & \textbf{4-gram} \\
        \midrule
        0.3 & 57.2 & 56.7 & 82.3 & 86.4 & 87.9 \\ 
        0.4 & 55.8 & 57.8 & 84.3 & 88.4 & 90.0 \\ 
        0.5 & 52.3 & 59.0 & 85.9 & 90.3 & 91.8 \\ 
        0.6 & 51.4 & 60.1 & 87.4 & 91.7 & 93.1 \\ 
        0.8 & 52.3 & 59.0 & 85.9 & 90.3 & 91.8 \\ 
        1.0 & 56.7 & 61.0 & 88.8 & 93.1 & 94.6 \\ 
        \bottomrule
    \end{tabular}
    }
    %\vspace{-5pt}
    \caption{Ablation studies on $\beta$. Too low $\beta$ can lead to unstable training, and too high $\beta$ constraint model from learning knowledge about how to distinguish hallucinations.}
    \label{tab:table_2}
\vspace{-5pt}
\end{table}

\begin{table*}[th]
    \centering
    \resizebox{0.85\textwidth}{!}{%
    \begin{tabular}{c|l|c|c|c|c|c}
        \toprule
        \textbf{POPE} & \textbf{Model} & \textbf{HA-DPO} & \textbf{Accuracy} & \textbf{Precision} & \textbf{F1 Score} & \textbf{Yes Ratio~(\%) } \\
        \midrule
        \multirow{6}{*}{\textbf{Random}} & \multirow{2}{*}{MiniGPT-4-LLama2-7B~\cite{DBLP:journals/corr/abs-2304-10592}} & \XSolidBrush & 51.13 & 50.57 & 67.13 & 98.66 \\
        & & \CheckmarkBold & 86.13 & 92.81 & 84.96 & 42.20 \\
        \cmidrule{2-7}
        & \multirow{2}{*}{InstructBLIP-13B~\cite{DBLP:journals/corr/abs-2305-06500}} & \XSolidBrush & 88.70 & 85.03 & {89.26} & 55.23 \\
        &  & \CheckmarkBold & {89.83} & \textbf{93.07} & {89.43} & {46.23} \\
        \cmidrule{2-7}
        & \multirow{2}{*}{LLaVA-1.5-7B~\cite{DBLP:journals/corr/abs-2310-03744}} & \XSolidBrush & 89.60 & 88.77  & {89.70} & \textbf{51.06} \\
        &  & \CheckmarkBold & \textbf{90.53} & {92.99}  & \textbf{90.25} & {47.13} \\
        \midrule
        \multirow{6}{*}{\textbf{Popular}}
        & \multirow{2}{*}{MiniGPT-4-LLama2-7B~\cite{DBLP:journals/corr/abs-2304-10592}} & \XSolidBrush  & 51.46 & 50.74 & 67.72 & 98.06 \\
        & & \CheckmarkBold & 79.50 & 80.20 & 79.25 & 48.83 \\
        \cmidrule{2-7}
        & \multirow{2}{*}{InstructBLIP-13B~\cite{DBLP:journals/corr/abs-2305-06500}} & \XSolidBrush  & 81.36 & 75.06  & 83.44 & 62.56 \\
        & & \CheckmarkBold & {85.76} & {85.55}  & {85.80} & \textbf{50.03} \\
        \cmidrule{2-7}
        & \multirow{2}{*}{LLaVA-1.5-7B~\cite{DBLP:journals/corr/abs-2310-03744}} & \XSolidBrush & {86.20} & 83.23 & {86.79} & 54.46 \\
        &  & \CheckmarkBold & \textbf{87.90} & \textbf{88.07}  & \textbf{87.81} & {49.76} \\
        \midrule
        \multirow{6}{*}{\textbf{Adversarial}}
        & \multirow{2}{*}{MiniGPT-4-LLama2-7B~\cite{DBLP:journals/corr/abs-2304-10592}} & \XSolidBrush  & 51.26 & 50.64  & 67.16 & 98.40 \\
        &  & \CheckmarkBold  & 75.66 & 74.36 & 76.29 & \textbf{52.66} \\
        \cmidrule{2-7}
        & \multirow{2}{*}{InstructBLIP-13B~\cite{DBLP:journals/corr/abs-2305-06500}} & \XSolidBrush  & 74.50 & 67.64  & 78.64 & 69.43 \\
        &  & \CheckmarkBold  & {80.70} & {77.72}  & {81.68} & {55.36} \\
        \cmidrule{2-7}
        & \multirow{2}{*}{LLaVA-1.5-7B~\cite{DBLP:journals/corr/abs-2310-03744}} & \XSolidBrush & {79.76} & {74.43}  & {81.75} & {60.90} \\
        &  & \CheckmarkBold & \textbf{81.46} & \textbf{77.99} & \textbf{82.54} & {56.20} \\
        \bottomrule
    \end{tabular}
    }
    \caption{Results on POPE Benchmark: HA-DPO significantly enhances the model's ability to discern hallucinatory objects in images.}
    \label{tab:table_3}
\vspace{-15pt}
\end{table*}

\noindent \textbf{Evaluation with POPE.}~%The POPE dataset is a mainstream evaluation dataset for hallucination in large multimodal models, with 9,000 questions across three types. The questions are confined to whether targets of fixed categories (80 COCO categories) exist in the image, with answers provided as Yes/No. Accuracy is calculated based on the model's answers and the ground-truth results. Although this can give some insight into the model's hallucination capabilities, it is firstly category-based and does not take into account object attributes or relative relationships.
%The POPE dataset is a mainstream evaluation dataset for hallucination evaluation in large multimodal models, featuring 9,000 questions in three types. It focuses on the presence of fixed categories (80 COCO categories) in images and provides Yes/No answers. The model's accuracy is determined by comparing its answers to the ground truth. %However, POPE is category-based and does not consider other hallucination types such as object attributes or relative relationships.
The POPE dataset, a mainstream dataset for hallucination evaluation in multimodal models, contains 9,000 questions of 3 types. POPE targets at object existence of fixed categories (80 COCO) in images, supplying Yes/No responses. The model's accuracy is benchmarked against the ground truth answer.

\noindent \textbf{Evaluation with SHR:}~%SHR evaluation contains 200 images from VG as the evaluation set~(val). Model is prompted to describe the given image in detail and the ratio of hallucinated sentences is reported as the main metric. The determination of hallucination and non-hallucination is made by GPT-4 based on the model's response and the corresponding VG annotations and human-annotated factual information for the current image.
%The SHR evaluation employs 200 VG images during its validation process, where the model is tasked to detailly describe each image. The main metric used is the proportion of hallucinated sentences. GPT-4 decides whether a sentence is hallucinated or not, using the model's output alongside VG annotations and human-annotated factual information of the present image.
The SHR evaluation uses 200 images from the VG dataset. During the evaluation, the model is required to provide detailed descriptions for these 200 images. SHR then measures the ratio of hallucinated sentences in these descriptions. GPT-4 determines if a sentence is hallucinated by comparing the model output with VG annotations and human-annotated factual information.

\subsection{{Implementation Details}}

\noindent\textbf{MiniGPT-4.} We fine-tune MiniGPT-4 model~\cite{DBLP:journals/corr/abs-2304-10592} parameters via LoRA~\cite{DBLP:conf/iclr/HuSWALWWC22}, fixing all but $q_{proj}$, $k_{proj}$ and $v_{proj}$. The LoRA's dimensionality (rank) is $64$ and $\alpha$ is 16. A cosine scheduler adjusts the learning rate at $1e^{-4}$ initially, with a warmup ratio of $0.03$ for $100$ rounds. The batch size is $1$, and HA-DPO's $\beta$ and $\lambda$ is set to $0.1$ and $0.5$. Training is performed on 8$\times$A100 GPUs takes 1--2 hours for 1K steps.

\noindent\textbf{InstructBLIP.} We fine-tune the InstructBLIP-13B model~\cite{DBLP:journals/corr/abs-2305-06500} parameters with LoRA, setting its rank to $64$ and $\alpha$ to 16. $q_{proj}$, $k_{proj}$, $v_{proj}$ in the language model are fine-tuned.  We use a learning rate of $4e^{-6}$ with a cosine learning rate schedulera a batch size of $1$, and set HA-DPO's $\beta$ as $0.1$ with $\lambda$ being $0$. Training runs on 8$\times$A100 GPUs for 1 epoch~(less than 1 hour).

\noindent\textbf{LLaVA-1.5.} Experiments on the LLaVA-1.5-7B model~\cite{DBLP:journals/corr/abs-2310-03744} involve fine-tuning all linear layers, using LoRA with a rank of $256$ and $\alpha$ of 128, following the original LLaVA-1.5 configuration. The learning rate, batch size, and hyperparameters $\beta$ and $\lambda$ is set to $2e^{-6}$, 16, 0.1, and 0 respectively, with learning rate adjusted by a cosine scheduler. Training is conducted using 8$\times$A100 GPUs for 1 epoch, which takes less than 1 hour.

\subsection{Ablation Studies}

% \subsubsection{Ablation on Style-consistent Control}

% We examine the role of data style-consistent control in preventing training instability. We fine-tune the MiniGPT4-LLaMA2 model with and without style-consistency control and evaluate on a subset of SHR. Instability is evaluated based on the degree of model degeneration, using n-gram fluency as an indicator, represented by $\frac{set(ngrams(s))}{len(ngram(s))}$, where $s$ stands for the image description generated by the model. High n-gram fluency indicates smoother sentences, whereas low values denote repetition and degeneration, which is the main indicator of failure in preference learning~\cite{DBLP:journals/corr/abs-2305-18290}. Table~\ref{tab:table_1} demonstrates that applying style-consistent control can ease training instability and reduce degeneration.

\subsubsection{Ablation on the Effect of Beta}
We conducted experiments on different hyper-parameter $\beta$ to verify their effects in HA-DPO. We carried out an experiment of MiniGPT4-LLaMA2 on a subset of SHR. Results are shown in Table~\ref{tab:table_2}. It can be observed that when $\beta$ is too small, HA-DPO training is unstable, and what the model mostly learns is noise rather than how to distinguish hallucinations. When $\beta$ is very large, the loss is more focused on constraining the consistency between the policy model and the reference model, which will also result in the model's inability to learn how to distinguish non-hallucination from hallucination.

\begin{figure*}[ht]
\begin{center}
	\includegraphics[width=\linewidth]{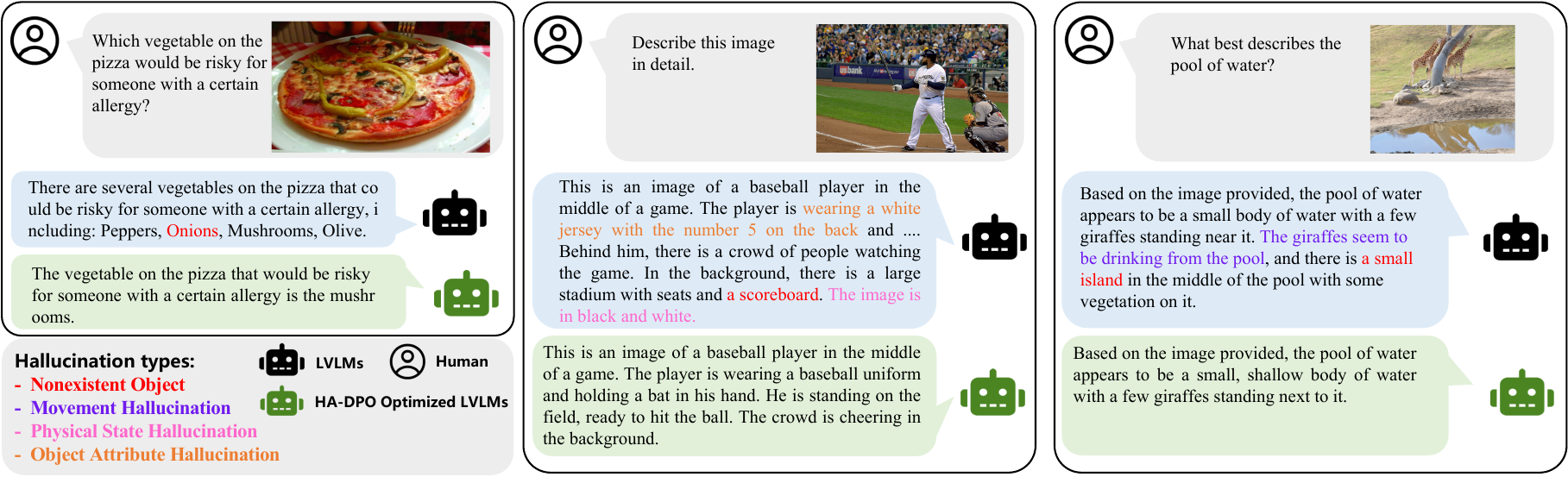}
\end{center}
\vspace{-8pt}
\caption{Comparison of model responses before and after our proposed hallucination elimination method.}
\label{fig:fig7}
\end{figure*}

\subsection{{Main Results}}

\subsubsection{Results on Hallucination Mitigation}
The phenomenon of hallucination is significantly reduced in models optimized with HA-DPO. We analyze this from both POPE and SHR perspectives.

\noindent\textbf{POPE.} The results of the POPE experiment are shown in Table~\ref{tab:table_3}. As can be seen, The phenomenon of hallucination in MiniGPT-4 is significantly reduced after optimization with HA-DPO. The accuracy metrics on the Random, Popular, and Adversarial sets improved by $35.0\%$, $28.04\%$, and $24.4\%$, respectively, and F1-Score improved by $17.83\%$, $11.53\%$, and $9.13\%$, respectively. LLaVA-1.5, due to the diversity of data in the supervised fine-tuning phase, has the least hallucination. After fine-tuning with HA-DPO, the performance of this model is further improved, reaching a new SOTA result of $90.25\%$ F1 score. Besides, HA-DPO optimized model achieves better performance on not only random set, but also more challenging set such as popular and adversarial set. Meanwhile, we compare HA-DPO with other hallucination mitigation methods, as shown in Table~\ref{tab:table_comp}. Results show that HA-DPO outperforms other competitive methods and achieves SOTA~(state-of-the-art) in POPE accuracy and F1 score. Notably, LRV and LLaVA-RLHF used 400K and 160K training data, respectively, while HA-DPO only used $2,000$ images and corresponding 16K pairs of positive and negative reply samples.

\begin{table}[t]
    \centering
    \resizebox{0.35\textwidth}{!}{
    \begin{tabular}{l|c|c}
        \toprule
        \textbf{Model} & \textbf{HA-DPO} & \textbf{SHR~$\downarrow$} \\
        \midrule
        \multirow{2}{*}{MiniGPT-4-LLama2-7B} & \XSolidBrush & 47.3  \\
        & \CheckmarkBold & \textbf{44.4} \\
        \midrule
        \multirow{2}{*}{InstructBLIP-13B} & \XSolidBrush & 51.2  \\
        & \CheckmarkBold & \textbf{49.1} \\
        \midrule
        \multirow{2}{*}{LLaVA-1.5-7B} & \XSolidBrush & 36.7  \\
        & \CheckmarkBold & \textbf{34.0} \\
        \bottomrule
    \end{tabular}
    }
    \caption{Hallucination evaluation results on SHR benchmark.}
    \label{tab:table_4}
\vspace{-5pt}
\end{table}

\noindent\textbf{SHR.} To more intuitively evaluate the model's hallucination at the sentence level, we use the SHR evaluation metric. As shown in Table~\ref{tab:table_4}, the original MiniGPT-4's SHR metric is $47.3\%$, meaning that nearly half of the sentences in its detailed descriptions of images exhibit hallucination, a rather alarming figure. After optimization with HA-DPO, the proportion of hallucinatory sentences decreased by about $3\%$.  An improvement of $2.1\%$ and $2.7\%$ are also achieved in InstructBLIP and LLaVA-1.5. However, we can see that the problem of LVLM hallucination remains severe.

\subsubsection{Results on General Performance Enhancement}

In order to verify the impact of HA-DPO's elimination of multimodal model hallucination on the model's general capabilities, we evaluated it on MME~\cite{DBLP:journals/corr/abs-2306-13394}, which aims to evaluate the general ability of LVLMs. The results, shown in Table~\ref{tab:table_5}, indicate that the MiniGPT-4 model optimized with HA-DPO significantly improved its MME metrics. The Perception metric increased from $726.72$ to $1051.41$, and Cognition increased from $169.64$ to $233.57$. As for InstructBLIP, an improvement of $71.32$ is also seen in Perception. The general capabilities of the model showed a clear improvement after the elimination of hallucinations, further demonstrating the necessity of eliminating hallucinations.

\subsubsection{Hallucination Elimination Examples}

In this section, we illustrate extensive hallucination elimination examples to further validate the effectiveness of HA-DPO. As shown in Figure~\ref{fig:fig7}, the application of the HA-DPO strategy significantly reduces multiple kinds of hallucinations, including but limited to object existence hallucination, object attribute hallucination, movement hallucination, and physical hallucination, etc. 

\begin{table}[t]
    \centering
    \resizebox{0.47\textwidth}{!}{%
    \begin{tabular}{l|c|c|c}
        \toprule
        \textbf{Model} & \textbf{HA-DPO} & \textbf{Perception} & \textbf{Cognition} \\
        \midrule
        \multirow{2}{*}{MiniGPT-4-LLama2-7B} & \XSolidBrush & 733.79 & 198.21  \\
        & \CheckmarkBold & \textbf{1092.18} & \textbf{234.28} \\
        \midrule
        \multirow{2}{*}{InstructBLIP-13B} & \XSolidBrush & 1344.91 & 232.50  \\
        & \CheckmarkBold & \textbf{1416.23} & \textbf{233.21} \\
        \midrule
        \multirow{2}{*}{LLaVA-1.5-7B} & \XSolidBrush & \textbf{1510.74} & \textbf{355.71}  \\
        & \CheckmarkBold & {1502.58} & {313.93} \\
        \bottomrule
    \end{tabular}
    }
    % \vspace{-5pt}
    \caption{Results on MME Benchmark. Recognition and cognition each represent two major capability dimensions in MME, examining recognition and perceptual reasoning ability, respectively.}
    \label{tab:table_5}
\vspace{-5pt}
\end{table}

\section{Conclusion}
\label{sec:conclusion}

% In this work, we introduced the Hallucination-Aware Delusional Perception Optimization (DPO) strategy and developed style-consistent hallucination data to enhance multimodal models. The refined models favor non-hallucinatory outputs and show significant generalization improvements. We also introduced the Sentence-level Hallucination Ratio (SHR), a new metric to directly quantify hallucination in model outputs. Our future work will focus on the application of these techniques in real-world scenarios. We aim to construct an evaluation system that can effectively assess and eliminate hallucinations, further validating our strategies and evaluation methods.

% In this work, we presented the Hallucination-Aware Direct Perception Optimization strategy and utilized style-consistent hallucination data to boost LVLMs. The improved models favor non-hallucinatory outputs, demonstrating enhanced generalization. Additionally, we introduced the Sentence-level Hallucination Ratio (SHR) for direct hallucination quantification in model outputs. In subsequent research, we aim to adapt these techniques to real-world scenarios, striving to establish a precise framework for effective hallucination identification and reduction.

In this work, we present the Hallucination-Aware Direct Perception Optimization strategy and utilize style-consistent hallucination data to enhance the performance of LVLMs. The improved models tend to generate non-hallucinatory outputs, demonstrating enhanced generalization capabilities. Additionally, we introduce the Sentence-level Hallucination Ratio (SHR) as a direct measure of hallucination in model outputs. In future research, we aim to adapt these techniques to real-world scenarios, with the goal of establishing a precise framework for effective hallucination identification and reduction.

%\newpage
{
    \small
    \bibliographystyle{ieeenat_fullname}
    \bibliography{main}
}
\clearpage
\setcounter{page}{1}
\maketitlesupplementary

\section{Dataset}
\textbf{Visual Genome (VG).} Visual Genome is a large-scale vision-language dataset that includes dense captions~\cite{DBLP:journals/ijcv/KrishnaZGJHKCKL17}. It contains over 100,000 densely annotated images, each averaging 21 objects, 18 attributes, and 18 object relationships. As the largest and densest dataset of image descriptions, objects, attributes, relationships, and question answers, VG bridges visual concepts with language. In our study, VG images are used to create both the hallucination training dataset and the SHR evaluation set.

\noindent\textbf{SHR Evaluation Set.} The SHR evaluation set, dedicated to evaluating the LVLM's hallucination at the sentence level, comprises 200 images from the Visual Genome. To ensure precise hallucination judgment, GPT-4 is provided with detailed annotations from Visual Genome and manually annotated factual information. There's no overlap between the SHR set images and those in the positive-negative data.

\section{Details of Hallucination Data Generation}

In this section, we delve into the specifics of the hallucination data generation process. This is illustrated through concrete examples at each of its three crucial stages: description generation, hallucination detection and correction, and style-consistency data augmentation.

\subsection{Description Generation}

We provide the images from Visual Genome dataset to LVLMs with the instruction ``Describe the image in detail.'' (as shown in Figure~\ref{fig:fig8}). For both hallucination dataset construction and SHR evaluation, generation parameters are configured as shown in Table~\ref{tab:supp_1}.

\begin{table}[h]
    \centering
    \begin{tabular}{c|c|c}
    \toprule
    \textbf{num\_beams} & \textbf{temperature} & \textbf{do\_sample} \\
    \midrule
    5 & 1.0 & False \\
    \bottomrule
    \end{tabular}
    \caption{Generation parameters.}
    \label{tab:supp_1}
\end{table}

\subsection{Hallucination Detection \& Correction}
We utilize GPT-4 to detect hallucinations in sentences produced by LVLMs, leveraging the detailed and rich image information provided by the annotations of the VG dataset. GPT-4 is prompted to identify hallucinated sentences and rectify these sentences. Any sentence deemed as a hallucination is preserved as a negative sample, while its corrected version is treated as a positive sample. An example is demonstrated in Figure~\ref{fig:fig9}.

\subsection{Style-consistency Data Augmentation}
Following hallucination detection and correction, we further prompt GPT to rewrite both the positive and negative samples. This ensures style-consistency between positive and negative samples. An instance of negative augmentation is displayed in Figure~\ref{fig:fig10} and an instance of positive augmentation is presented in Figure~\ref{fig:fig11}.

%\begin{table}[t]
%    \centering
%    \begin{subtable}[t]{0.495\linewidth}
%        \centering  
%        \footnotesize
%        \begin{tabular}{c|c|c}
%        \toprule
        %\diagbox{\textbf{H}}{\textbf{G}} & 
%        \textbf{TP} & \textbf{non-hal} \\
%        \midrule
%        \textbf{H} & 127 & 15 \\
%        \textbf{No. H} & 16 & 107 \\
%        \bottomrule
%        \end{tabular}
%        \caption{w/o factual information.}
%        \label{tab1:shr}
%    \end{subtable}
%    \begin{subtable}[t]{0.495\linewidth}
%        \centering
%        \footnotesize
%        \begin{tabular}{c|c|c}
%        \toprule
%        \diagbox{\textbf{H}}{\textbf{G}} & \textbf{hal} & \textbf{non-hal} \\
%        \midrule
%        \textbf{H} & 131 & 11 \\
%        \textbf{No. H} & 3 & 120 \\
%        \bottomrule
%        \end{tabular}
%        \caption{w factual information.}
%        \label{tab2:shr}
%    \end{subtable}
%    \caption{Human judgment vs GPT-4 judgment. "H" stands for Human judgment and "G" stands for GPT-4 judgment. With manually added factual information, the GPT-4 judgment is more accurate and aligned with human judgment. }
%    \label{tab:shr}
%\end{table}

\begin{table}[t]
    \centering
    \begin{subtable}[t]{0.495\linewidth}
        \footnotesize
        \begin{tabular}{c|c|c}
        \toprule
        & \textbf{P} & \textbf{R} \\
        \midrule
        \textbf{H} & 88.81\% & 89.43\% \\
        \textbf{C} & 86.99\% & 87.70\% \\
        \bottomrule
        \end{tabular}
        \caption{w/o factual information.}
        \label{tab3:shr}
    \end{subtable}
    \begin{subtable}[t]{0.495\linewidth}
        \footnotesize
        \begin{tabular}{c|c|c}
        \toprule
        & \textbf{P} & \textbf{R} \\
        \midrule
        \textbf{H} & 97.76\% & 92.25\% \\
        \textbf{C} & 92.60\% & 97.56\% \\
        \bottomrule
        \end{tabular}
        \caption{w factual information.}
        \label{tab4:shr}
    \end{subtable}
    \caption{Precision and Recall of GPT-4 judgment. "P" stands for precision and "R" stands for recall. "H" stands for hallucination and "C" stands for correct. The precision and recall of GPT-4 judgment is much improved with human-annotated factual information.}
    \label{tab:shr}
\end{table}

\section{Details of SHR evaluation}
\label{supp:sec9}
\subsection{Evaluation Example}

In the SHR evaluation, GPT-4 classifies each sentence in the model response as \textbf{hallucination} or \textbf{correct}. The SHR is then computed as the proportion of hallucinated sentences to total sentences. Consult Figure~\ref{fig:fig12} for illustration.

\subsection{Factual-assisted Evaluation}

During the SHR evaluation, some inaccuracies in GPT-4's judgments were observed due to insufficient VG annotations. To improve the judgment accuracy of GPT-4, we manually supplement each image with extra factual information. Therefore, incorrect GPT-4 judgments are corrected with this additional information. Next, to validate the effect of additional factual information, we meticulously check judgments of GPT-4 and responses of three LVLMs~(MiniGPT-4, LLaVA-1.5, and InstructBLIP) on 20 images~(about 250 sentences), both with and without this factual information. %Results (as shown in Table~\ref{tab1:shr} and \ref{tab2:shr}) clearly indicate a significant decrease in false positives. 
Tables \ref{tab:shr} present improved precision and recall with the help of additional factual information. Overall, human-annotated factual annotation enhanced GPT-4's judgment accuracy from 88.30\% to 95.84\%.

\begin{figure}[t]
    \centering
    \includegraphics[width=0.48\textwidth]{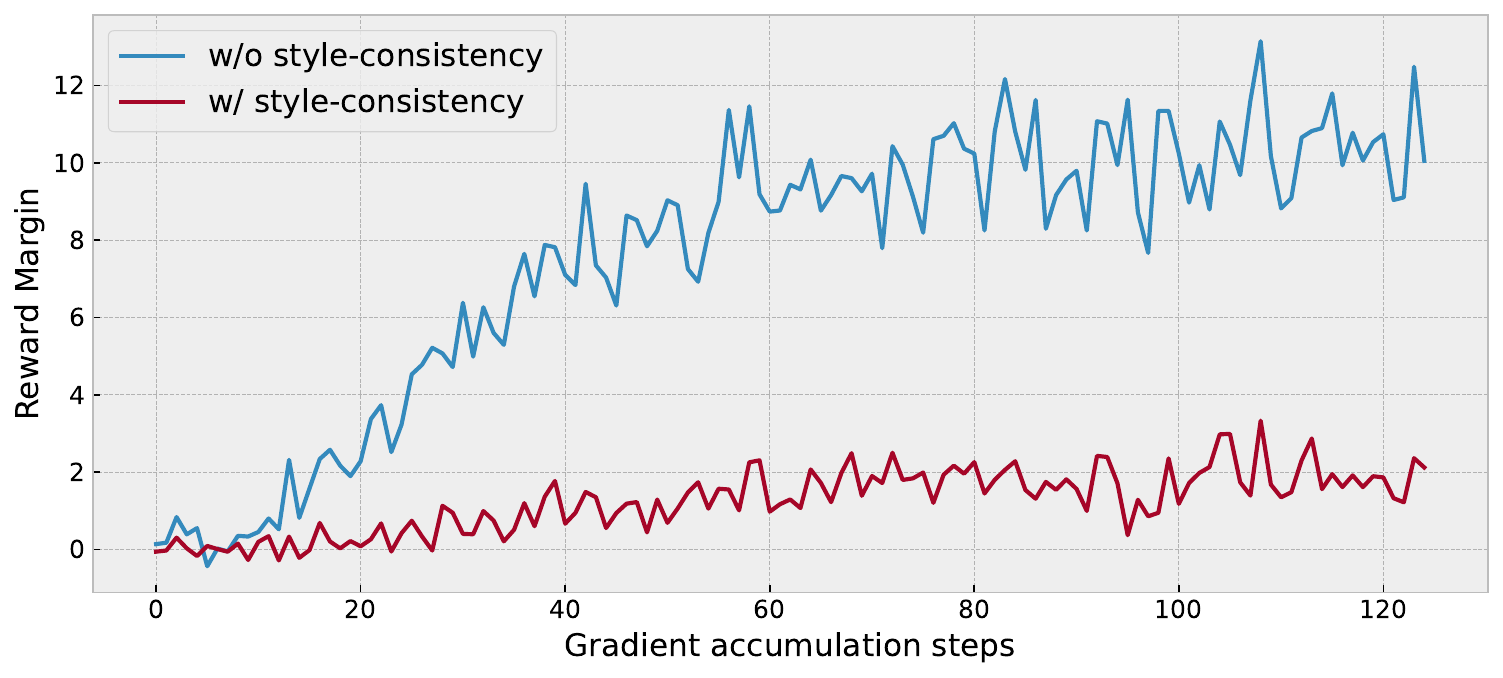}
    \caption{Comparison of gradient value between training with and without style-consistency data. Gradient changing using style-consistent data is much smoother.}
    \label{fig:fig5}
\end{figure}

\begin{table*}[th]
    \centering
    \resizebox{0.9\textwidth}{!}{%
    \begin{tabular}{l|cc|cc|cc}
        \toprule
        \multirow{2}{*}{\textbf{Method}} & \multicolumn{2} {c|}{\textbf{Random}} & \multicolumn{2} {c|}{\textbf{Popular}}& \multicolumn{2} {c}{\textbf{Adversarial}} \\
        & \textbf{Accuracy} & \textbf{F1 score} & \textbf{Accuracy} & \textbf{F1 score} & \textbf{Accuracy} & \textbf{F1 score} \\
        \midrule
        LRV~\cite{DBLP:journals/corr/abs-2306-14565} & 86.00 & 88.00 & 73.00 & 79.00 & 65.00 & 73.00 \\
        LLaVA-RLHF~\cite{DBLP:journals/corr/abs-2309-14525} & 84.80 & 83.30 & 83.90 & 81.80 & \textbf{82.30} & 80.50 \\
        \cmidrule{1-7}
        {LLaVA-1.5-7B} & 89.60 & 89.70 & \underline{86.20} & \underline{86.79} & 79.76 & \underline{81.75} \\
        {InstructBLIP-13B} & 88.70 & 89.26 & 81.36 & 83.44 & 74.50 & 78.64 \\
        {MiniGPT4-LLaMA2-7B} & 51.13 & 67.13 & 51.46 & 67.72 & 51.26 & 67.16 \\
        \cmidrule{1-7}
        \rowcolor{gray!20} LLaVA-1.5-7B \scriptsize \textbf{\textit{w HA-DPO}} & \textbf{90.53} & \textbf{90.25} & \textbf{87.90} & \textbf{87.81} & \underline{81.46} & \textbf{82.54} \\
        \rowcolor{gray!20} InstructBLIP-13B \scriptsize \textbf{\textit{w HA-DPO}} & \underline{89.83} & \underline{89.43} & 85.76 & 85.80 & 80.70 & 81.68 \\
        \rowcolor{gray!20} MiniGPT4-LLaMA2-7B \scriptsize \textbf{\textit{w HA-DPO}} & 86.13 & 84.96 & 79.50 & 79.25 & 75.66 & 76.29 \\
        \bottomrule
    \end{tabular}
    }
    \caption{Results comparisons with other hallucination mitigation methods on POPE Benchmark. HA-DPO outperforms other competitive methods and achieves SOTA~(state-of-the-art) in POPE accuracy and F1 score. Bolded denotes the best score and underline denotes the second best score.}
    \label{tab:table_comp}
%\vspace{-15pt}
\end{table*}

\section{Additional Style-consistency Analysis}
\label{supp:sec10}

To further demonstrate the effect of style-consistent control, we quantitatively examine the role of data style-consistent control in preventing training instability. Specifically, MiniGPT4-LLaMA2 model is fine-tuned with and without style-consistency control and evaluated on a subset of SHR. Instability is evaluated based on the degree of model degeneration, using n-gram fluency as an indicator, represented by $\frac{set(ngrams(s))}{len(ngram(s))}$, where $s$ stands for the image description generated by the model. High n-gram fluency indicates smoother sentences, whereas low values denote repetition and degeneration, which is the main indicator of failure in preference learning~\cite{DBLP:journals/corr/abs-2305-18290}. Table~\ref{tab:table_1} demonstrates that applying style-consistent control can ease training instability and reduce degeneration.

To allow readers to better understand the importance of style-consistency, we further provide some quantitative analysis regarding style consistency here. As shown in Eq~\ref{eq3}, we mentioned that the lack of style consistency could lead to a disfunctioning of the weight factor, thereby causing extremely unstable gradients in preference learning. To further validate this conclusion, we demonstrate the gradient variates with and without style consistency in Figure~\ref{fig:fig5}. It can be observed that optimizing gradient on a style-consistent dataset is much more stable. %Meanwhile, we show the degenerate phenomenon, which is the main indicator of model failure in DPO, in Figure~\ref{fig:fig6}. Without style-consistent control, n-grams fail quickly, which indicates that training can easily diverge and fall into degeneration.

\begin{table}[t]
    \footnotesize
    \centering
    \resizebox{0.48\textwidth}{!}{%
    \begin{tabular}{c|c|c|c|c|c}
        \toprule
        \textbf{$\beta$} & \textbf{Style Consistency} & \textbf{1-gram} & \textbf{2-gram} & \textbf{3-gram} & \textbf{4-gram} \\
        \midrule
        \multirow{2}{*}{{0.1}} & \XSolidBrush & 17.9 & 23.9 & 25.2 & 25.9 \\ 
        & \Checkmark & 56.8 & 83.5 & 88.1 & 90.0 \\
        \midrule
        \multirow{2}{*}{{0.3}} & \XSolidBrush & 40.8 & 58.0 & 61.1 & 62.3 \\
        & \Checkmark & 58.7 & 87.0 & 91.7 & 93.4 \\
        \midrule
        \multirow{2}{*}{{0.5}} & \XSolidBrush & 47.3 & 69.1 & 73.0 & 74.6 \\
        & \Checkmark & 59.4 & 87.8 & 92.4 & 93.9 \\
        \midrule
        \multirow{2}{*}{{1.0}} & \XSolidBrush & 57.2 & 85.3 & 90.1 & 91.8 \\
        & \Checkmark & 60.1 & 89.1 & 93.8 & 95.4 \\
        \midrule
        \multicolumn{2}{c|}{\textbf{MiniGPT-4}} & 60.0 & 91.7 & 96.6 & 98.0 \\
        \bottomrule
    \end{tabular}
    }
    %\vspace{-5pt}

    \caption{Ablation studies on on the effect of style-consistency control in stabilizing training and mitigating degeneration.}
    \label{tab:table_1}
\vspace{-5pt}
\end{table}

\begin{figure*}[t]
   \centering
   \includegraphics[width=0.999\textwidth]{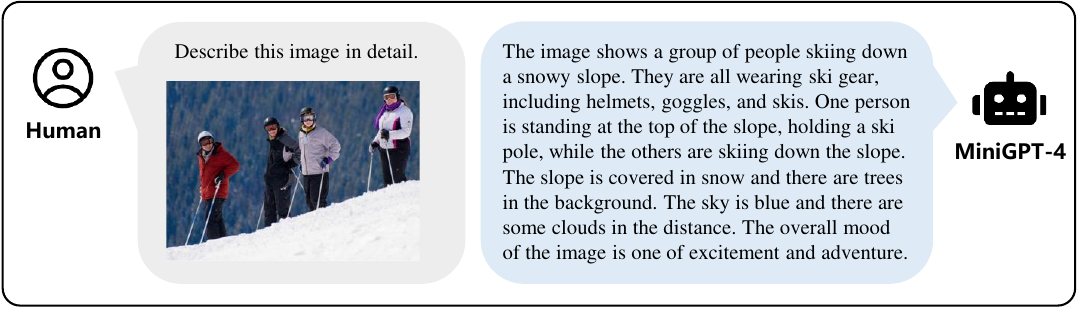}
   \caption{Detailed Image Description Generation Using LVLMs.}
   \vspace{-1pt}
   \label{fig:fig8}
\end{figure*}

\section{Comparison with other hallucination mitigation methods}
\label{supp:sec11}

In this section, we compare our proposed method with other hallucination mitigation methods, results are shown in Table~\ref{tab:table_comp}. Results show that HA-DPO outperforms other competitive methods and achieves SOTA~(state-of-the-art) in POPE accuracy. Notably, LRV used $400,000$ training data, while our hallucination elimination only used $2,000$ images and corresponding $6,000$ pairs of positive and negative reply samples.

To clarify, the decrease in HA-DPO's POPE recall is not an indicator of model degradation. POPE Recall merely considers the correctness of "Yes" responses and overlooks "No" answers. For instance, models like MiniGPT-4, favoring "Yes" outputs, can attain near 100\% Recall, but with most "No" responses being wrong and accuracy is just 51\%. Therefore, Accuracy and F1 score are holistic measures for model hallucination in POPE.

\begin{figure*}[t]
   \centering
   \includegraphics[width=0.8\textwidth]{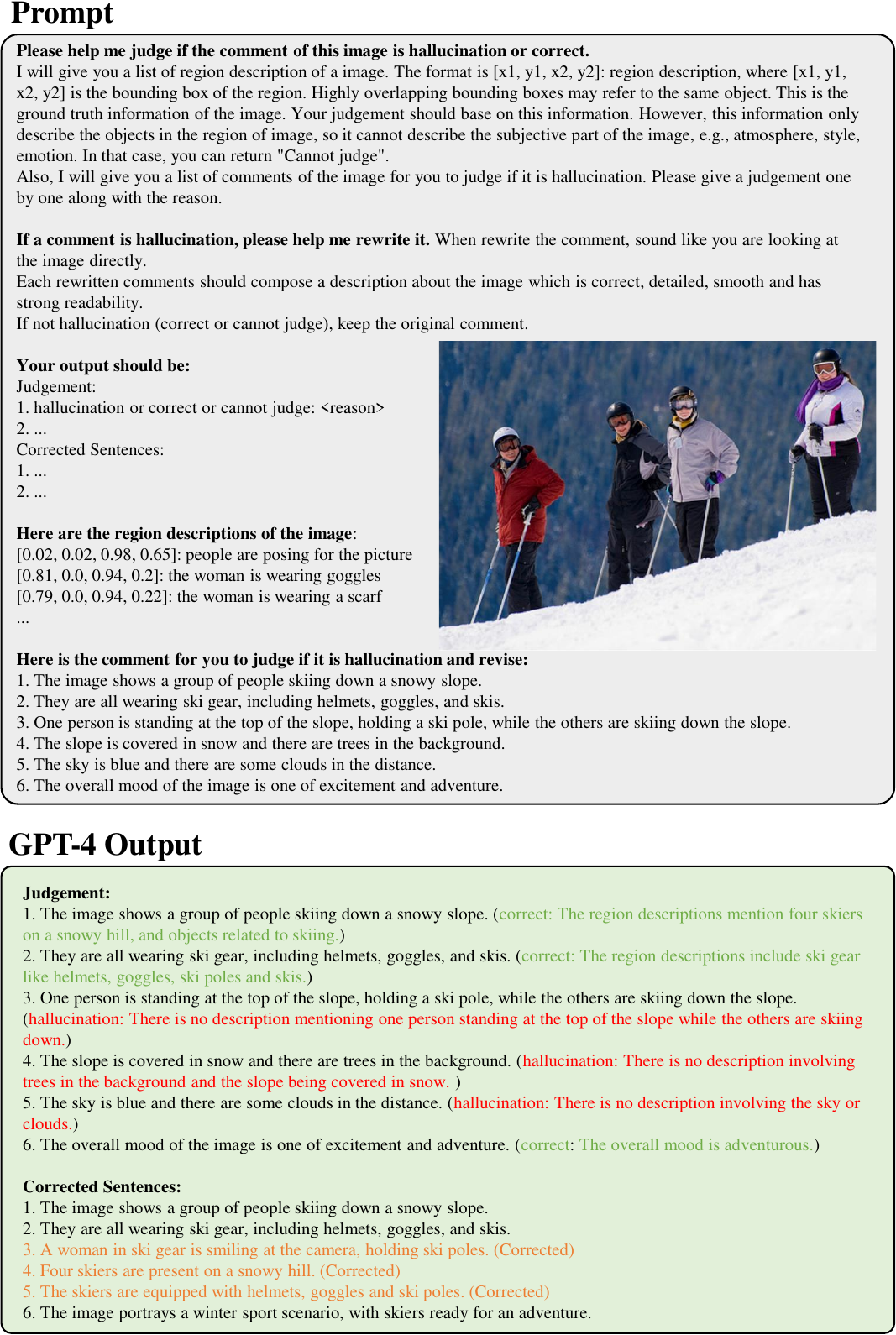}
   \caption{Hallucination Detection and Correction in LVLM-Generated Sentences.}
   \label{fig:fig9}
\end{figure*}

\begin{figure*}[t]
   \centering
   \includegraphics[width=0.9\textwidth]{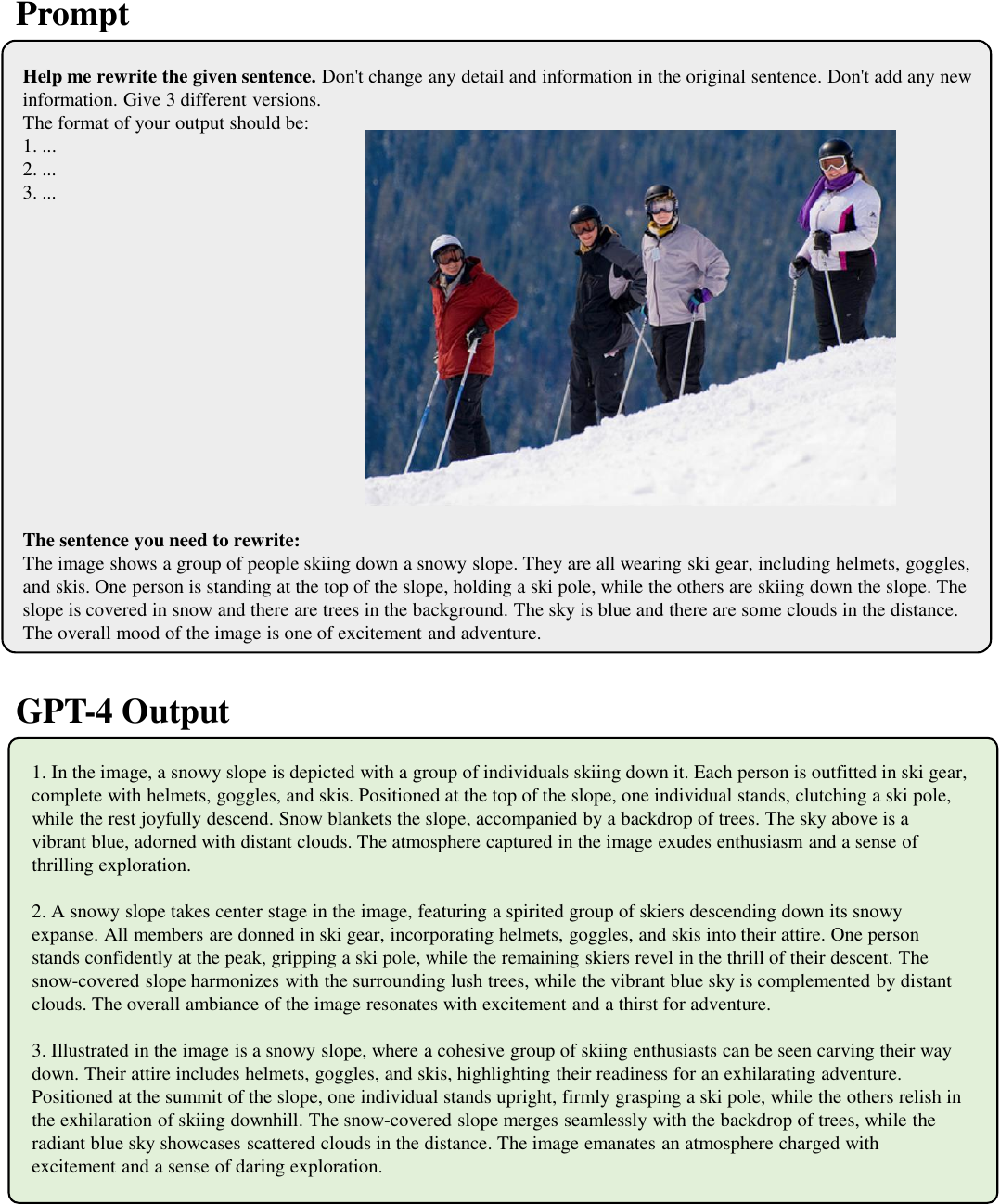}
   \caption{Style-Consistent Data Augmentation: Negative Sample (Hallucinated).}
   \label{fig:fig10}
\end{figure*}

\begin{figure*}[t]
   \centering
   \includegraphics[width=0.9\textwidth]{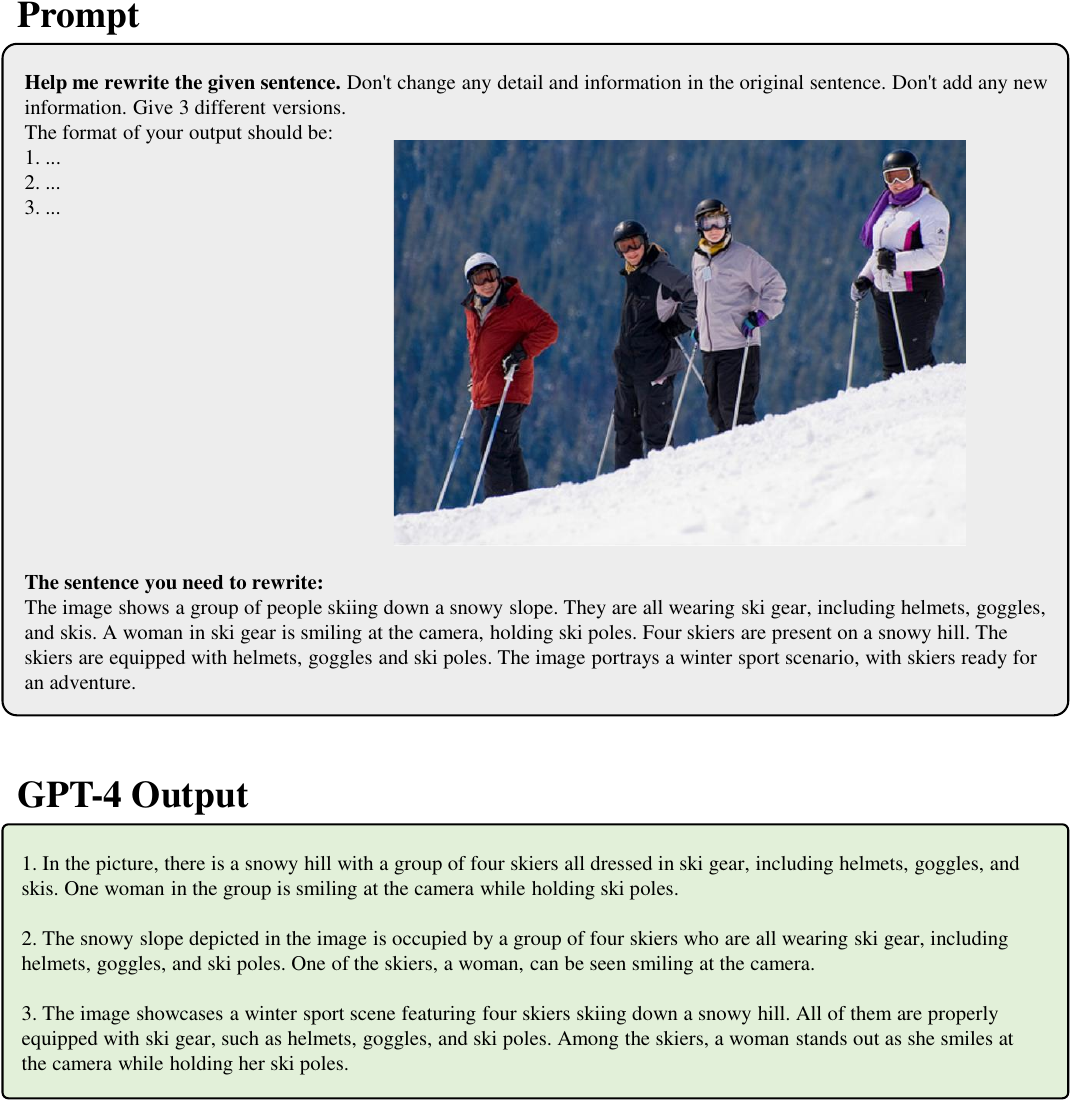}
   \caption{Style-Consistent Data Augmentation: Positive Sample (Non-hallucinated).}
   \label{fig:fig11}
\end{figure*}

\begin{figure*}[t]
   \centering
   \includegraphics[width=0.9\textwidth]{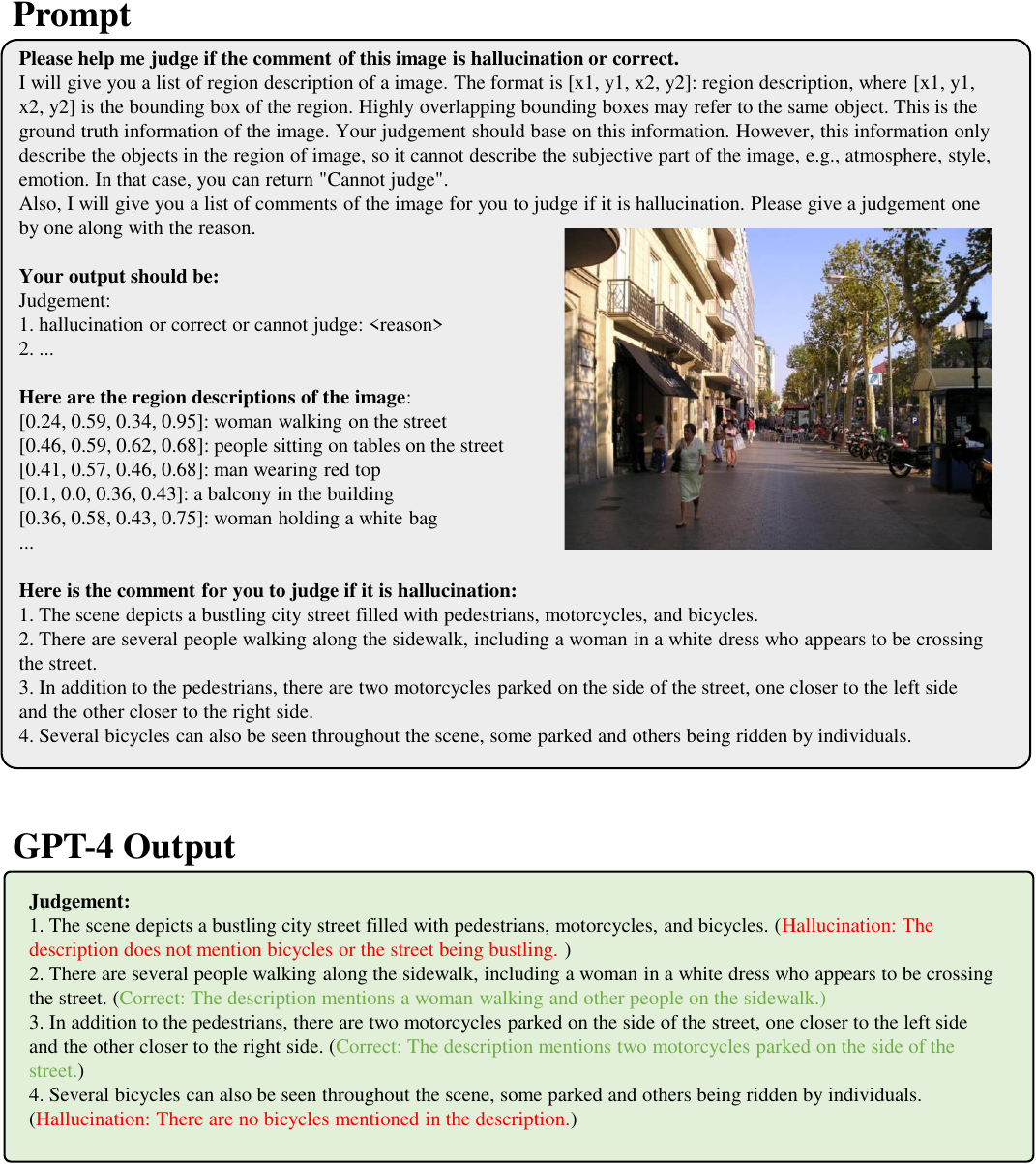}
   \caption{Illustration of Sentence Hallucination Ratio (SHR) Evaluation.}
   \label{fig:fig12}
\end{figure*}

\begin{figure*}[t]
   \centering
   \includegraphics[width=0.85\textwidth]{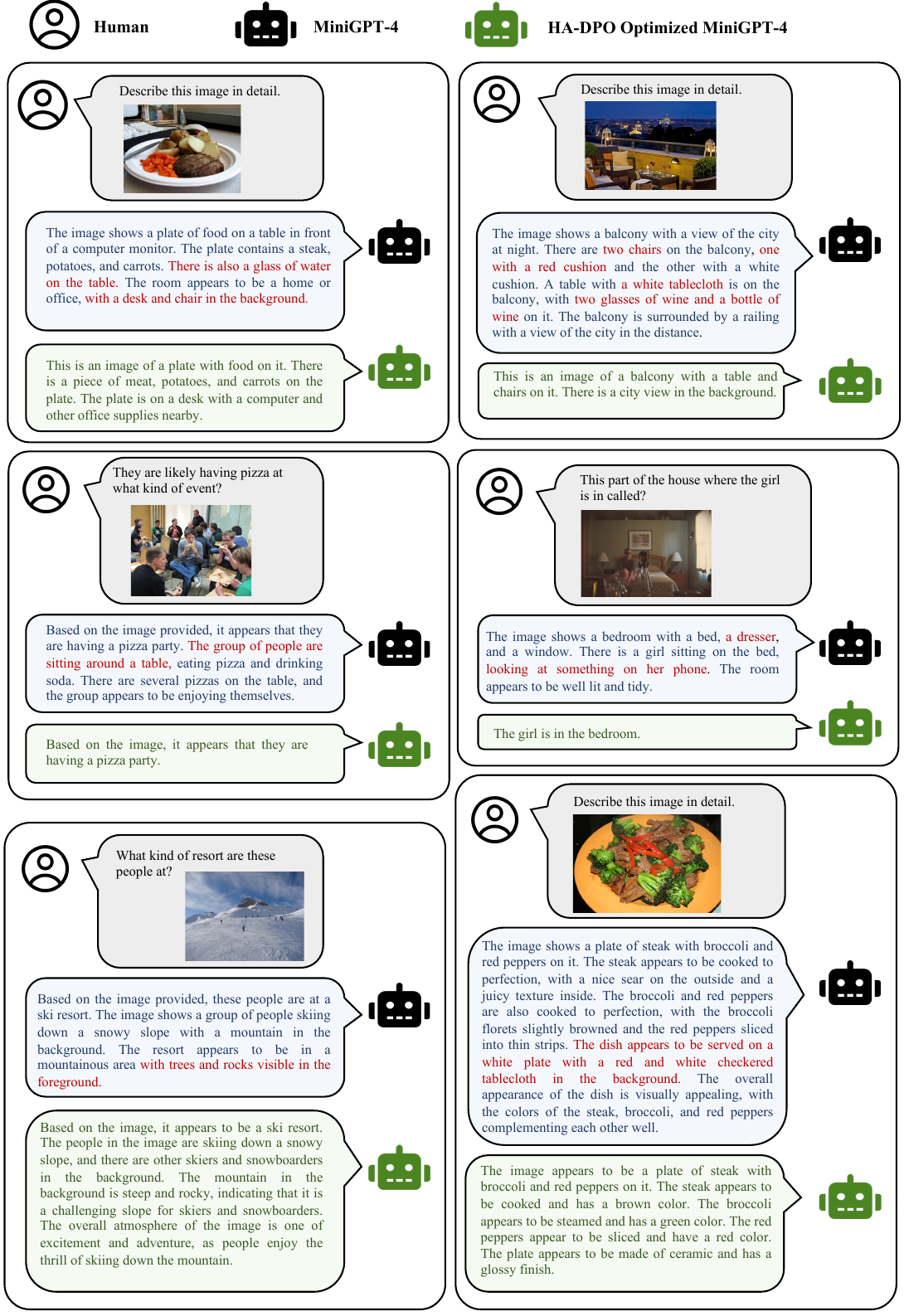}
   \caption{High-quality examples from MiniGPT4-LLaMA2-7B optimized by HA-DPO.}
   \label{fig:fig13}
\end{figure*}

\begin{figure*}[t]
   \centering
   \includegraphics[width=0.86\textwidth]{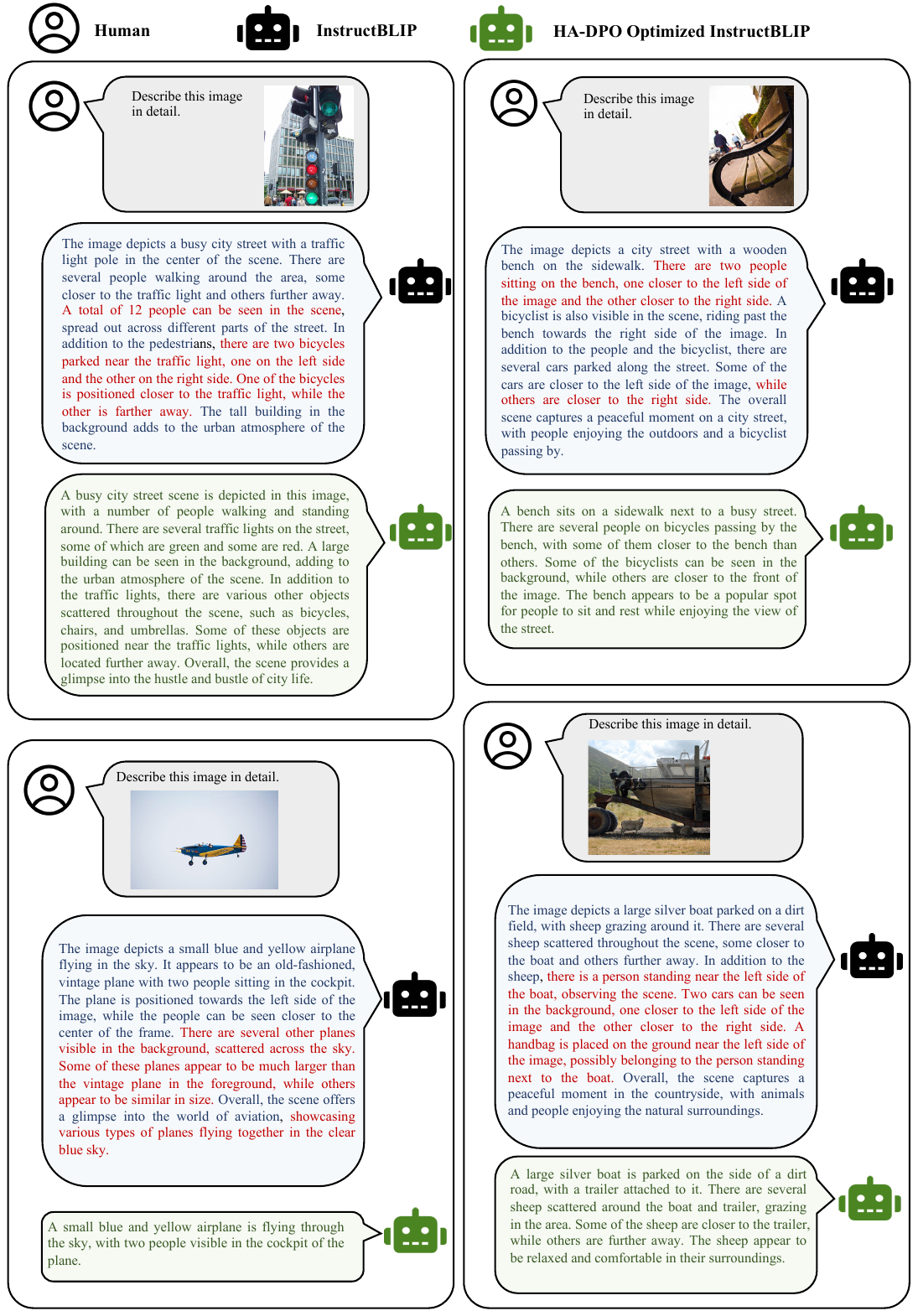}
   \caption{High-quality examples from InstructBLIP-13B optimized by HA-DPO.}
   \label{fig:fig14}
\end{figure*}

\begin{figure*}[t]
   \centering
   \includegraphics[width=0.86\textwidth]{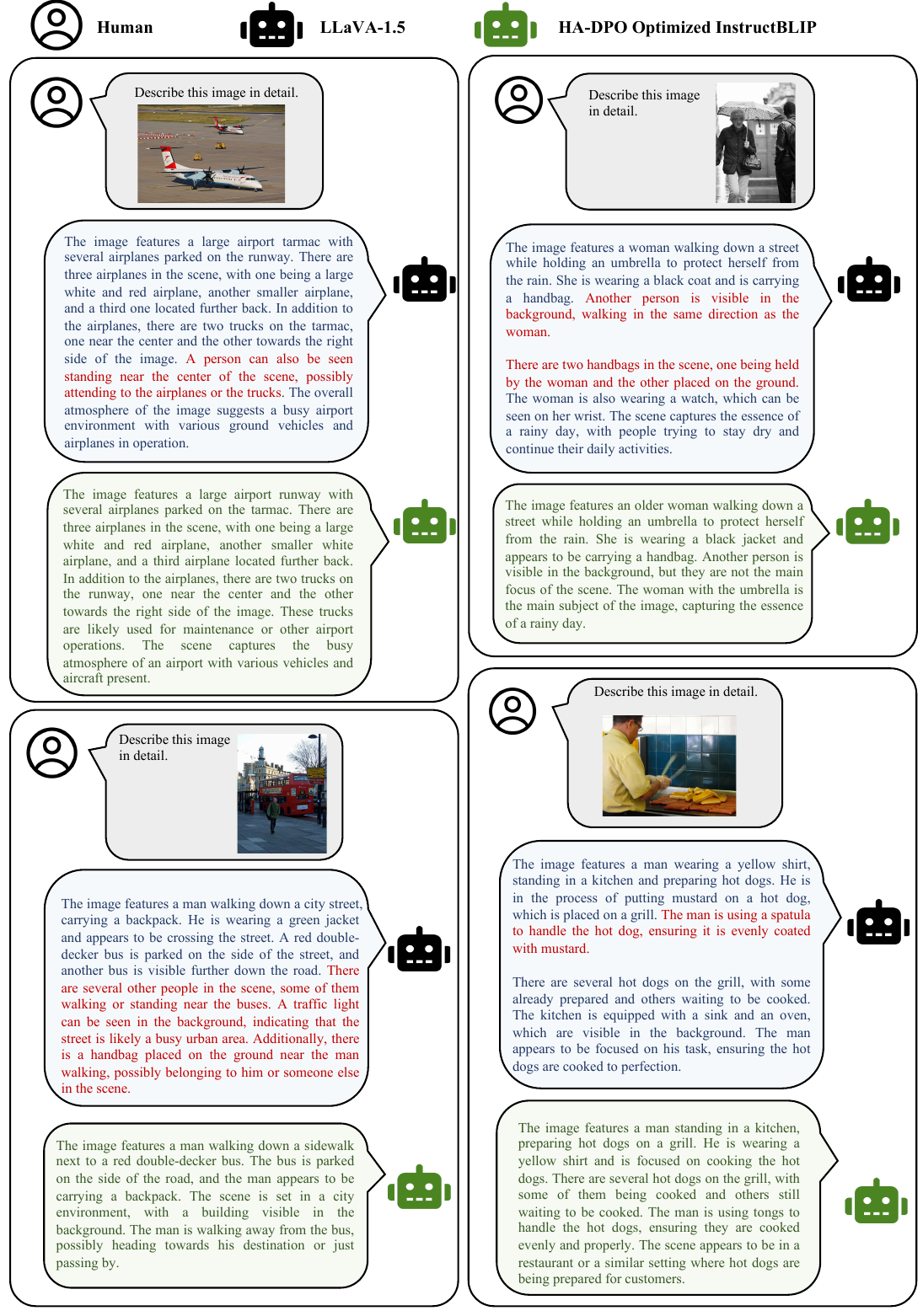}
   \caption{High-quality examples from LLaVA-1.5-7B optimized by HA-DPO.}
   \label{fig:fig15}
\end{figure*}

\section{Quality Examples}

Figure~\ref{fig:fig13}, Figure~\ref{fig:fig14} and Figure~\ref{fig:fig15}  present hallucination-eliminated examples generated by the MiniGPT4-LLaMA2-7B, InstructBLIP-13 and LLaVA-1.5, respectively. Models optimized using HA-DPO produce significantly less hallucinated content in both visual question-answering and image description tasks.

\subsection{Efficacy and Potential of HA-DPO}

According to the POPE and SHR evaluation results, we observe that the HA-DPO method can learn to output without hallucination bias during training, thereby providing more precise results and reducing the phenomenon of hallucination in large vision-language models. Given the effectiveness and simplicity of the HA-DPO method, we envision HA-DPO as a specialized stage following supervised fine-tuning, capable of rectifying biases in hallucination and potentially other dimensions. This would enable the model to produce outputs that are more realistic and in line with human expectations.

%\begin{table}[t]
%    \centering
%    \resizebox{0.35\textwidth}{!}{
%    \begin{tabular}{l|c|c}
%        \toprule
%        \textbf{Model} & \textbf{HA-DPO} & \textbf{SHR~$\downarrow$} \\
%        \midrule
%        \multirow{2}{*}{LLaVA-1.5} & \XSolidBrush & 46.9  \\
%        & \CheckmarkBold & 43.2~\textcolor{red}{(-3.7)} \\
%        \bottomrule
%    \end{tabular}
%    }
    % \vspace{-5pt}
%    \caption{Results of LLaVa-1.5 Model on the SHR Evaluation Set.}
%    \label{tab:supp_res_table2}
%\end{table}

\end{document}